\documentclass{article}

\usepackage{microtype}
\usepackage{graphicx}
\usepackage{subfigure}
\usepackage{booktabs} 

\usepackage{hyperref}

\usepackage[accepted]{icml2025}

\usepackage{amsmath}
\usepackage{amssymb}
\usepackage{mathtools}
\usepackage{amsthm}

\usepackage[capitalize,noabbrev]{cleveref}

\theoremstyle{plain}

\theoremstyle{definition}

\theoremstyle{remark}

\usepackage[textsize=tiny]{todonotes}

\usepackage{multirow}

\icmltitlerunning{Neural Scaling Laws Rooted in the Data Distribution}

\begin{document}

\twocolumn[
\icmltitle{Neural Scaling Laws Rooted in the Data Distribution}



\icmlsetsymbol{equal}{*}

\begin{icmlauthorlist}
\icmlauthor{Ari Brill}{independent}
\end{icmlauthorlist}

\icmlaffiliation{independent}{Independent}

\icmlcorrespondingauthor{Ari Brill}{aryeh.brill@gmail.com}

\icmlkeywords{deep learning, scaling laws, data manifold, percolation, emergence}

\vskip 0.3in
]



\printAffiliationsAndNotice{}  

\begin{abstract}
Deep neural networks exhibit empirical neural scaling laws, with error decreasing as a power law with increasing model or data size, across a wide variety of architectures, tasks, and datasets. This universality suggests that scaling laws may result from general properties of natural learning tasks. We develop a mathematical model intended to describe natural datasets using percolation theory. Two distinct criticality regimes emerge, each yielding optimal power-law neural scaling laws. These regimes, corresponding to power-law-distributed discrete subtasks and a dominant data manifold, can be associated with previously proposed theories of neural scaling, thereby grounding and unifying prior works. We test the theory by training regression models on toy datasets derived from percolation theory simulations. We suggest directions for quantitatively predicting language model scaling.
\end{abstract}

\section{Introduction}

Deep neural networks leverage massive amounts of computation to achieve superb performance on real-world tasks. In particular, neural scaling laws are the empirical finding that deep neural networks, including large language models (LLMs), perform predictably better as the model or dataset size increases, with the test error dropping as a power law over many orders of magnitude \cite{hestness2017deep, kaplan2020scaling, hoffmann2022training, bachmann2024scaling}. Neural scaling appears to be a universal phenomenon, with power-law exponents often of order $\sim0.1$ observed across a wide variety of architectures, tasks, and datasets \cite{henighan2020scaling, ghorbani2021scaling, jones2021scaling, alabdulmohsin2022revisiting, alabdulmohsin2024getting}.

It is unclear why neural scaling laws exist. One influential theoretical approach treats neural networks as approximating a continuous function defined on a manifold with low intrinsic dimension relative to the input space \cite{bahri2021explaining, sharma2022}. Increasing the model or dataset size allows for increasingly fine piecewise function approximation, producing power-law scaling with an exponent inversely proportional to the data manifold dimension. 

On the other hand, deep neural networks do more than approximate functions. They learn representations, constructing a hierarchy of abstract explanatory features that reflect the data's underlying structure \citep{bengio2013representation}. If a neural network achieves good performance by learning good features, it stands to reason that a scaled-up network that achieves even better performance may do so by learning more or better features. Indeed, works in mechanistic interpretability have empirically studied how scaling laws may connect to specific model behaviors and circuits  \cite{hernandez2022scaling, olsson2022context, michaud2024quantization}.

Several models have been proposed connecting neural scaling laws to power-law feature distributions. \citet{hutter2021learning} studied a toy model of learning, showing that a dataset of Zipf-distributed discrete features can yield power-law data scaling with a range of exponents. In kernel regression, power-law kernel spectra can produce power-law scaling laws \citep{spigler2020asymptotic, bordelon2020spectrum, bahri2021explaining}. \citet{maloney2022solvable} considered a generative data model based on a set of latent features with power-law spectral structure, and computed the resulting scaling laws in the context of a random feature model. Notably, \citet{michaud2024quantization} conjectured that natural prediction problems involve discrete subtasks or quanta, which can be ordered into a \textit{Q Sequence} by frequency of usefulness. By assuming that the use frequencies are power-law-distributed and models learn quanta in the order of the Q Sequence, power-law neural scaling laws can be derived.

These prior works show that power-law dataset structure can produce power-law neural scaling laws, but they do not explain why this structure would emerge across disparate domains. Unifying manifold-approximation and feature-learning theories of power-law scaling, and understanding how data distributions bound scaling exponents, have been identified as key research questions to resolve foundational challenges for assuring the alignment and safety of LLMs \citep{anwar2024foundational}. 

In this work, we propose a model of emergent power-law dataset structure that yields power-law neural scaling laws. We make two key assumptions meant to describe natural datasets, \textit{context-dependent target function} and \textit{general-purpose learning}, discussed in Sec.~\ref{sec:assumptions}. In Sec.~\ref{sec:percolation_model}, we introduce percolation theory and use it to translate these assumptions into a mathematical model of dataset structure. 

Two scaling regimes emerge that integrate and recontextualize previously proposed theoretical models of neural scaling laws. Below a critical threshold, the data distribution consists of a power-law distribution of subtasks, which we identify with the quanta proposed by \citet{michaud2024quantization}. Each quantum corresponds to a low-dimensional structure in data space. Above the critical threshold, a single structure dominates the data distribution, and we identify this regime with the manifold-approximation model proposed by \citet{sharma2022}.

In Sec.~\ref{sec:scaling_laws_derivation}, we derive theoretical scaling laws in model and data size, and perform experimental tests in Sec.~\ref{sec:experiments}. We discuss implications of the theory in Sec.~\ref{sec:discussion}, highlighting progress toward quantitatively predicting LLM scaling laws.

\section{Definitions and Assumptions}\label{sec:assumptions}

\subsection{Data Space}

We consider supervised regression with inputs $X \subseteq \mathbb{R}^{d'}$, target labels $Y \subseteq \mathbb{R}$, and mean squared error loss $\mathcal{L}$. The inputs and labels are sampled uniformly at random.

Real-world data often have physical invariances reflecting underlying symmetries of the physical generating process \cite{lin2017does, roberts2021ai}. For example, image data are subject to locality and translation invariance. Encoding such invariances into the model can improve sample complexity and reduce the data's effective dimension \citep{tahmasebi2023exact}. We accordingly define the \textit{data space} $\mathcal{X}$ with dimension $d \leq d'$ such that each axis represents one of the $d$ contingent, empirical degrees of freedom remaining after accounting for any symmetries or invariances. The domain of each dimension of $\mathcal{X}$ is the domain of its corresponding input degree of freedom.

In practice, continuous inputs can be distinguished only above a precision cutoff. This might be set by sensor resolution, human perception, or floating-point numerical precision. The number of distinct values along each data space dimension determines a characteristic length scale $L$. We approximate $\mathcal{X}$ as a hypercubic lattice with $L^d$ discrete elements, and discretize the labels $Y$ similarly. We denote discrete data-space elements as $x \in \mathcal{X}$, with corresponding labels $y \in Y$. For convenience, we'll write $f(x)$ for a function evaluated at any point corresponding to $x$ in the $(d' - d)$-dimensional manifold in the input space $X$.

\subsection{Context-Dependent Target Function}\label{sec:hof}

Modeling real-world data often requires different behaviors in different contexts. A language model might compose a sonnet when prompted with Shakespeare, write Python code given a software specification, and generate moves for a grandmaster-level chess game. In general, a useful target function could as well correspond to an accumulation of disparate behaviors as to a unified concept \citep{minsky1988society}.

We model a context-dependent target function as a higher-order target function $HOF$ that generates first-order functions, which then return the target labels. That is, $HOF$ generates first-order functions $f_x: \mathcal{X} \to Y$ such that $HOF(x) = f_x$ and $f_x(x) = y$. Each of the functions $f_x$ is assumed to be Lipschitz continuous on its domain, following \citet{sharma2022}.

Factorizing the target function this way splits the learning task into two parts. The functions $f_x$ determine the modeling difficulty of subtasks, while $HOF$ controls the difficulty of generalizing among subtasks. For example, a natural language dataset comprising multiple languages might be expected to have an $HOF$ with a greater rate of change than one composed of a single language.

\subsection{General-Purpose Learning}

A target function corresponding to a real-world task has latent structure rooted in the real, outside world. Reflecting reality, this structure is arbitrarily complex, contingent, and \textit{a priori} unknown. Reality's apparently unbounded complexity is one reason why AI systems applying general, scalable learning algorithms have outperformed methods reliant on built-in human knowledge \citep{sutton2019bitter}. The input representation used by a general-purpose learning system without built-in task-specific knowledge is essentially arbitrary, since by definition it's unrelated to the target function's extrinsic structure.

We assume a general-purpose learning task. Because the input representation is arbitrary, we'll model the data distribution statistically by assuming that the input representation is random. We elaborate on the conceptual underpinnings of the general-purpose learning assumption and the kinds of datasets to which it's meant to apply in Appendix~\ref{appendix:general_purpose_learning}.  

\section{Percolation Model}\label{sec:percolation_model}

\subsection{Setup}\label{sec:model_setup}

We draw a \textit{bond} between each pair of adjacent data-space elements $x_i, x_j \in \mathcal{X}$ if they require interchangeable functional behavior, that is, if $f_i(x_i) = f_j(x_i)$ and $f_i(x_j) = f_j(x_j)$. Let $p$ be the fraction of adjacent pairs connected by a bond. A group of connected elements forms a cluster. An element with no bonds is \textit{out of distribution}, because its function can't be learned via generalization, only memorized. Elements with bonds are \textit{in distribution}.  

Due to general-purpose learning, the bond pattern's projection in data space can be modeled as random. We analyze the dataset's overall statistical properties by considering a hypercubic lattice with elements occupied at random with probability $p$. We'll do so using percolation theory, which has been applied in fields including physics, materials science, network theory, biology, hydrology, and geochemistry \cite{sahimi1994applications}. Sec.~\ref{sec:percolation_theory} reviews foundational results in percolation theory, which can be found in an introductory textbook, such as \citet{stauffer1994introduction}.

\subsection{Percolation Theory}\label{sec:percolation_theory}

Percolation theory concerns the statistical and structural properties of clusters of randomly occupied units on a lattice or network. These systems exhibit phase transitions about a critical occupation probability $p_c$, called the percolation threshold. When $p < p_c$, clusters are finite and disconnected, while for $p \ge p_c$, the system \textit{percolates}: a so-called infinite cluster emerges that scales with the size of the system, its size going to infinity in an infinite lattice. The numerical value of $p_c$ depends on details of the system.

\begin{figure*}
    \centering
    \subfigure[$p = 0.1 \ll p_c$]{
        \centering
        \includegraphics[width=0.25\textwidth, trim=0.5cm 0.5cm 0.5cm 0.5cm, clip]{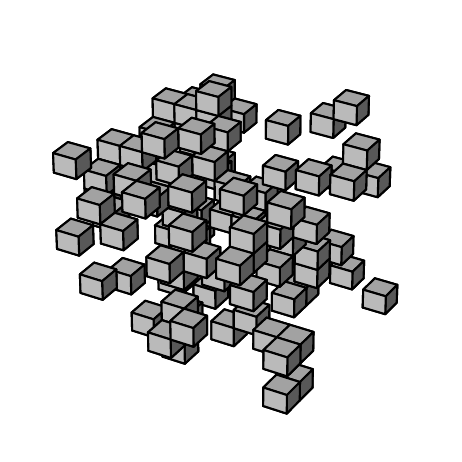}
    }
    \hfill
    \subfigure[$p = 0.3116 = p_c$]{
        \centering
        \includegraphics[width=0.25\textwidth, trim=0.5cm 0.5cm 0.5cm 0.5cm, clip]{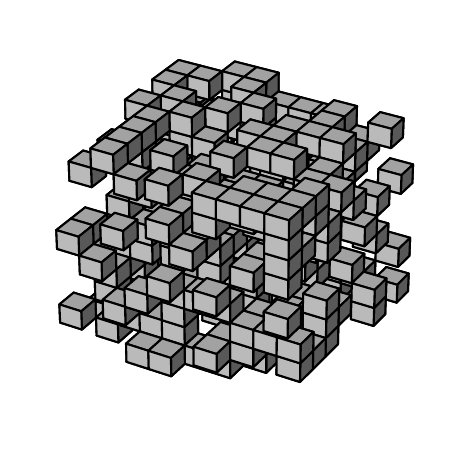}
    }
    \hfill
    \subfigure[$p = 0.6 \gg p_c$]{
        \centering
        \includegraphics[width=0.25\textwidth, trim=0.5cm 0.5cm 0.5cm 0.5cm, clip]{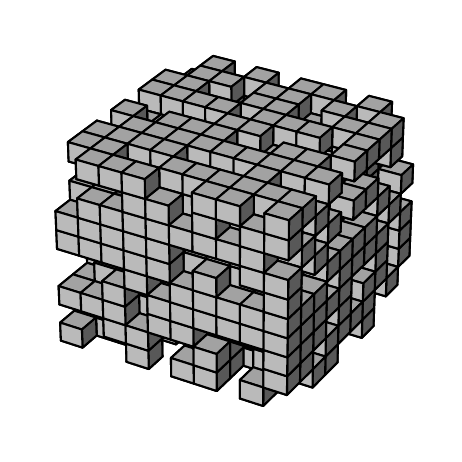}
    }
    \caption{Visualization of site percolation on a $10 \times 10 \times 10$ lattice.}
    \label{fig:percolation_visualization}
\end{figure*}

The randomly occupied units of percolation can be either lattice sites (site percolation) or bonds between neighboring sites (bond percolation). In either case, a cluster consists of a group of occupied units connected by a chain of nearest-neighbor relations, and a cluster's size $s$ is its total number of units. Fig.~\ref{fig:percolation_visualization} illustrates site percolation on a 3D lattice. Bond and site percolation are closely related, and all results here are valid in either model\footnote{Sec~\ref{sec:model_setup} described bonds between lattice sites. In the equivalent site percolation description, data-space elements are in-distribution with probability $p$, and groups of adjacent in-distribution elements form same-function clusters.}. Following \citet{stauffer1994introduction}, we'll use the language of site percolation.

On a large $d$-dimensional hypercubic lattice of size $L$, the number of finite clusters of size $s$ per lattice site, $n_s$, can be approximated for large $s$ as

\begin{equation}\label{eq:ns}
    n_s \propto s^{-\tau} e^{-as}, 
\end{equation}

where $a \propto |p - p_c|^{1/\sigma}$ and $\tau$ and $\sigma$ are universal critical exponents that depend only on $d$. At the percolation threshold $p = p_c$, the cluster size distribution is a power law, $n_s \propto s^{-\tau}$. Away from $p_c$, the size distribution is cut off exponentially, with most occupied sites instead belonging to the infinite cluster when $p > p_c$.

Clusters have nontrivial geometric structure. Finite clusters are fractal objects characterized by an intrinsic dimension $D < d$. Specifically, if an $s$-cluster's linear extent is described by its radius of gyration $R_s$, then $R_s \propto s^{1/D}$. At the percolation threshold, the incipient infinite cluster is also a fractal object with $R_s \propto s^{1/D}$, transitioning to Euclidean geometry with $R_s \propto s^{1/d}$ for $p \gg p_c$.

A realistic data space is likely high-dimensional. Percolation on a Bethe lattice, an infinite tree in which every site has the same number of edges, provides an accurate model of percolation on a lattice with any $d \geq 6$.  In this case, $\tau = 5/2$, $\sigma = 1/2$, and $D = 4$.

\section{Neural Scaling Laws}\label{sec:scaling_laws_derivation}

\subsection{Emergent Quanta at the Percolation Threshold}\label{sec:feature_distribution}

At the percolation threshold $p = p_c$, the probability that a site belongs to a cluster of size $s$ is given by

\begin{equation}
    p(s) = \frac{n_s s}{\int_1^\infty n_s s ds} = \frac{s^{-\tau}s}{\int_1^\infty s^{-\tau} s ds} = (\tau - 2)s^{-(\tau - 1)}.
\end{equation}

The probability that a site belongs to a cluster of size $s$ or larger is then given by

\begin{equation}\label{eq:cluster_cdf}
    P(s) = \int_s^\infty (\tau - 2)s'^{-(\tau - 1)} = s^{-(\tau - 2)}.
\end{equation}

We assign each cluster a rank $k$ based on its size, with  $k = 1$ indicating the largest cluster, $k = 2$ the next largest, and so on. To do so, we use the fact that the rank distribution is directly proportional to the cumulative distribution function, $k(s) \propto P(s)$ \cite{newman2005power}. By inverting Eq.~\ref{eq:cluster_cdf}, we obtain a Zipf power-law distribution

\begin{equation}\label{eq:s_k_relation}
    s \propto k^{-\frac{1}{\tau - 2}}.
\end{equation}

We show in Appendix~\ref{appendix:loss_reduction} that each cluster contributes to the baseline loss approximately in proportion to that cluster's size, $\Delta \mathcal{L}_s \propto s$. We have therefore derived a \textit{Q Sequence}, as proposed by \citet{michaud2024quantization}. Each cluster's function defines a ``quantum'' of model behavior to learn. The functions (quanta) form an ordered sequence in which fully learning the $k^\mathrm{th}$ quantum reduces the loss such that

\begin{equation}\label{eq:loss_k}
    -\Delta \mathcal{L}_k \propto k^{-(\alpha + 1)},
\end{equation}

with exponent $\alpha > 0$. Equating Eqs.~\ref{eq:s_k_relation} and \ref{eq:loss_k} gives $\alpha = (3 - \tau)/(\tau - 2)$. For $d \ge 6$, $\tau = 5/2$, yielding $\alpha = 1$. 

Following \citet{michaud2024quantization}, throughout this work we will sometimes use the terms ``quantum'' and ``Q Sequence'' to refer to a function defined on a discrete cluster and to the sequence of quanta ordered by cluster size, respectively.

\subsection{Model Scaling}\label{sec:model_scaling}

We now study scaling in model size. We consider resolution-limited scaling \citep{bahri2021explaining}, taking the dataset size and computational resources to be effectively infinite. We assume that the model is nonparametric and characterized by $N$ independent effective degrees of freedom (DOF). Empirical scaling laws are typically reported in terms of parameter count $P$. The correct way to convert between DOF and parameters may be architecture-dependent. We discuss this issue in Appendix~\ref{appendix:dofandparameters}. We report scaling laws in terms of $N$ and leave open the conversion to $P$ for future investigation. 

In nonparameteric regression and density estimation, the error with respect to DOF (or data samples) scales exponentially in the input dimension $d$, $\epsilon \propto N^{-1/\mathcal{O}(d)}$ \citep{wasserman2006all}. This results from the curse of dimensionality: $\mathcal{O}\left((1/l)^d\right)$ hypercubes are required to piecewise approximate a function by slicing space into hypercubes with side length $l$. Furthermore, if the data lie on a low-dimensional manifold of intrinsic dimension $D < d$, deep neural networks can achieve much more efficient $c/D$ scaling, for a constant $c$ \citep{chen2019, sharma2022, bahri2021explaining}. For a piecewise constant model, $c \ge 2$, and for a piecewise linear model such as a ReLU neural network, $c \ge 4$. Equality holds for generic Lipschitz continuous functions, with faster scaling possible for non-generically simple functions \cite{sharma2022}.

The data distribution at $p_c$ consists of power-law-distributed clusters each with intrinsic dimension $D$. From Eq.~\ref{eq:loss_k}, each cluster contributes to the loss in proportion to its size. If $n_k$ of the model's $N$ effective DOF are used to approximate the function of the rank-$k$ cluster, then that cluster's loss contribution further scales as $n_k^{-c/D}$. The sum of cluster losses with DOF allocated optimally, accounting for both factors, gives the overall scaling law. 

The predicted model scaling law is derived in Appendix~\ref{appendix:model_size_scaling}. Fig.~\ref{fig:theoretical_model_scaling} shows how it changes as $c/D$ varies relative to $\alpha$. Intuitively, we can think about the loss-minimizing model as allocating DOF to model those cluster functions that on the margin most reduce the loss. At first, this is the largest cluster only. As that cluster becomes better modeled, DOF are also assigned to the next-largest cluster, and so on, following the Q Sequence. The optimal DOF allocation is

\begin{figure}
    \centering
    \includegraphics[width=0.9\linewidth]{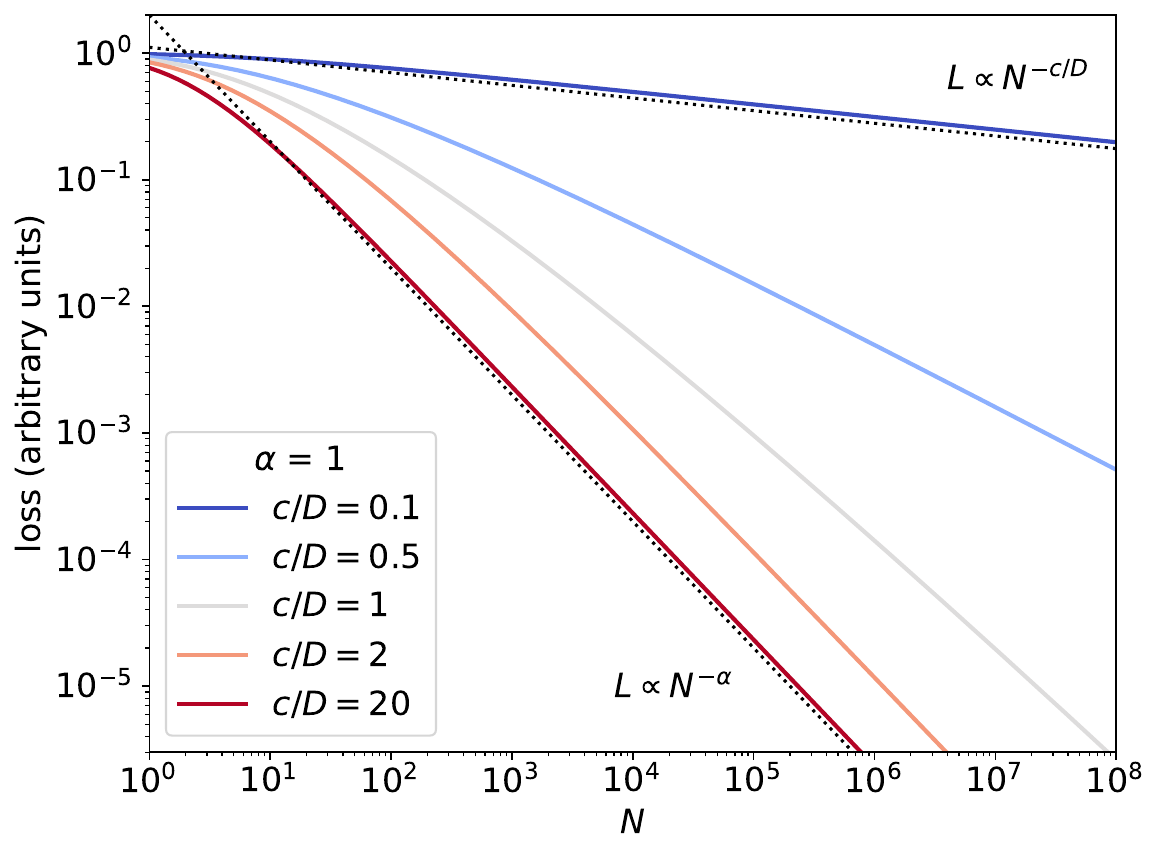}
    \caption{Theoretical scaling law as a function of model DOF $N$, for $\alpha = 1$ and various values of $c/D$.}
    \label{fig:theoretical_model_scaling}
\end{figure}

\begin{equation}\label{eq:dof_allocation}
    n_k = \begin{cases} 
      ak^{b - 1} & k < k_\mathrm{br}\\
      0 & k \geq k_\mathrm{br}.
   \end{cases}
\end{equation}

where $a$ is a normalization factor, the allocation exponent is

$$
    b = 1 - \frac{\alpha + 1}{c/D + 1} = \frac{c/D - \alpha}{c/D + 1},
$$

and the cluster rank where a break occurs is

$$
    k_\mathrm{br} \approx \begin{cases}
        N & c/D \gg \alpha,~ c/D \gtrsim 1\\
        \left(\ln{\frac{N}{\ln{N}}}\right)^{-1}N & c/D \approx \alpha \mathrm{~or~} c/D \ll 1,~ \alpha \ll 1\\
        \left(|b|N\right)^{1/(1 + |b|)} & c/D \ll \alpha,~ \alpha \gtrsim 1.
    \end{cases}
$$

The predicted loss is 

\begin{equation}\label{eq:model_scaling}
    \mathcal{L} \propto \left(\frac{N}{k_\mathrm{br}} + \frac{1}{\alpha}\right)k_\mathrm{br}^{-\alpha}.
\end{equation}

In the limiting case where $c/D \gg \alpha$ and $c/D \gtrsim 1$, each cluster function is quickly modeled well. The loss is determined by the number of clusters learned, scaling as

\begin{equation}
    \mathcal{L} \sim \left(1 + \frac{1}{\alpha}\right) N^{-\alpha}.
\end{equation}

When $c/D \ll \alpha$ and $\alpha \gtrsim 1$, clusters are modeled slowly, and only the functions for a handful of the largest clusters can be learned. For large $N$, the loss scales as

\begin{equation}
     \mathcal{L} \sim \left(\frac{\alpha}{c/D + 1}\right)^{-(c/D + 1)}  \left(1 + \frac{1}{\alpha} N^{-\frac{\alpha}{1 + \alpha}} \right) N^{-c/D}.
\end{equation}

In the critical regime, $\alpha = 1$ (Sec~\ref{sec:feature_distribution}). For large clusters when $d \ge 6$, $D = 4$. For ReLU activations, $c = 4$, so $c/D = 1$. A ReLU neural network or equivalent is thus the minimally powerful function approximator for which model scaling begins to transition from the manifold approximation to the Q Sequence regime for a data distribution at criticality.

\subsection{Data Scaling}

We now consider resolution-limited scaling in dataset size $\mathcal{D}$ (i.e. number of training examples). We idealize the model as an optimal approximator, with the finite dataset size bottlenecking performance. While before DOF were distributed to optimize performance, each cluster's expected number of data points is now fixed by its size.

A cluster's size affects its loss in three (opposing) ways. First, larger size means more test data points to model, increasing the loss. Second, larger size means more training data points for modeling its function, decreasing the loss in accordance with manifold approximation scaling. Third, larger size increases the probability that the cluster is represented in the randomly sampled training data, decreasing the loss due to the chance it might not be learned at all. The total loss is the sum of each cluster's loss.

Intuitively, if the cluster functions are relatively easy to model, then the first and third considerations dominate, and the data scaling law will reflect the cluster size distribution. Conversely, if the functions are relatively hard to model, then the second consideration is most important, and the scaling law will correspond to manifold approximation.

\begin{figure}
    \centering
    \includegraphics[width=0.9\linewidth]{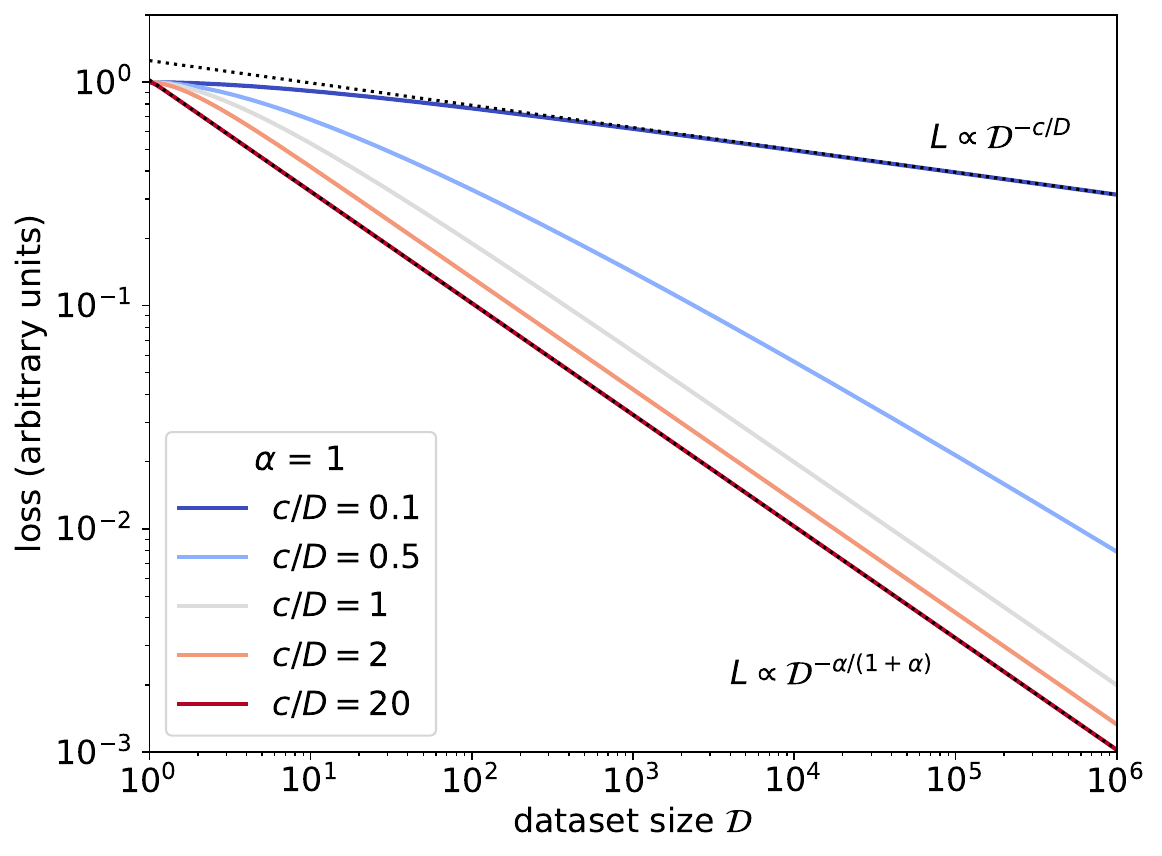}
    \caption{Theoretical scaling law as a function of dataset size $\mathcal{D}$, for $\alpha = 1$ and various values of $c/D$.}
    \label{fig:theoretical_data_scaling}
\end{figure}

The predicted data scaling law is derived in Appendix~\ref{appendix:data_scaling}. Fig.~\ref{fig:theoretical_data_scaling} shows how it changes as $c/D$ varies relative to $\alpha$. We obtain, up to an overall constant factor,

\begin{equation}\label{eq:data_scaling}
    \mathcal{L} \approx \frac{\alpha^{-\alpha/(1 + \alpha) - 1}}{1 - \frac{\alpha/(1 + \alpha)}{c/D}}\mathcal{D}^{-\alpha/(1 + \alpha)} + \frac{\alpha^{-c/D - 1}}{1 - \frac{c/D}{\alpha/(1 + \alpha)}}\mathcal{D}^{-c/D},
\end{equation}

assuming that $c/D \ne \alpha/(1 + \alpha)$. In the limiting case where $c/D \gg \alpha/(1 + \alpha)$, the first term dominates and we obtain $\mathcal{L} \propto \mathcal{D}^{-\alpha/(1 + \alpha)}$. In the limiting case where $\alpha/(1 + \alpha) \gg c/D$, the second term dominates and we obtain $\mathcal{L} \propto \mathcal{D}^{-c/D}$. If $c/D = \alpha/(1 + \alpha)$, we obtain

\begin{equation}
    \mathcal{L} \approx \alpha^{-c/D - 1}\left(1 + \frac{c}{D}\log\alpha + \frac{c}{D}\log\mathcal{D} \right) \mathcal{D}^{-c/D}.
\end{equation}

\subsection{Subcritical and Supercritical Scaling}

In the subcritical regime $p < p_c$, the cluster size distribution has an exponential cutoff for large $s$ (Eq.~\ref{eq:ns}), but the power-law exponent is the same. Scaling should thus be similar to the critical regime $p = p_c$. A detailed analysis is left for future work. The supercritical regime $p > p_c$ is qualitatively different. The infinite cluster dominates the data distribution, and a single function rather than a Q Sequence best describes the learning task. This regime appears to correspond to the manifold approximation model proposed by \citet{sharma2022}. The intrinsic manifold dimension $D$ can be identified with the infinite cluster's fractal dimension, which is the same as for finite clusters when $p = p_c$ and becomes Euclidean, $D = d$, when $p \gg p_c$.

\subsection{Scaling Regimes}

The derived scaling laws have two main free parameters: $p$, which relates to the dataset's degree of context-dependence, and the manifold approximation constant $c$, which represents the model's ability to fit complex functions. In the subcritical regime, for high-dimensional data, $\alpha$ and $D$ are fixed by percolation theory, varying only for $d < 6$. In the supercritical regime, $D$ is determined by the dimension of the infinite cluster, which depends on $p$.

\begin{figure}[tb]
    \centering
    \includegraphics[width=0.7\linewidth]{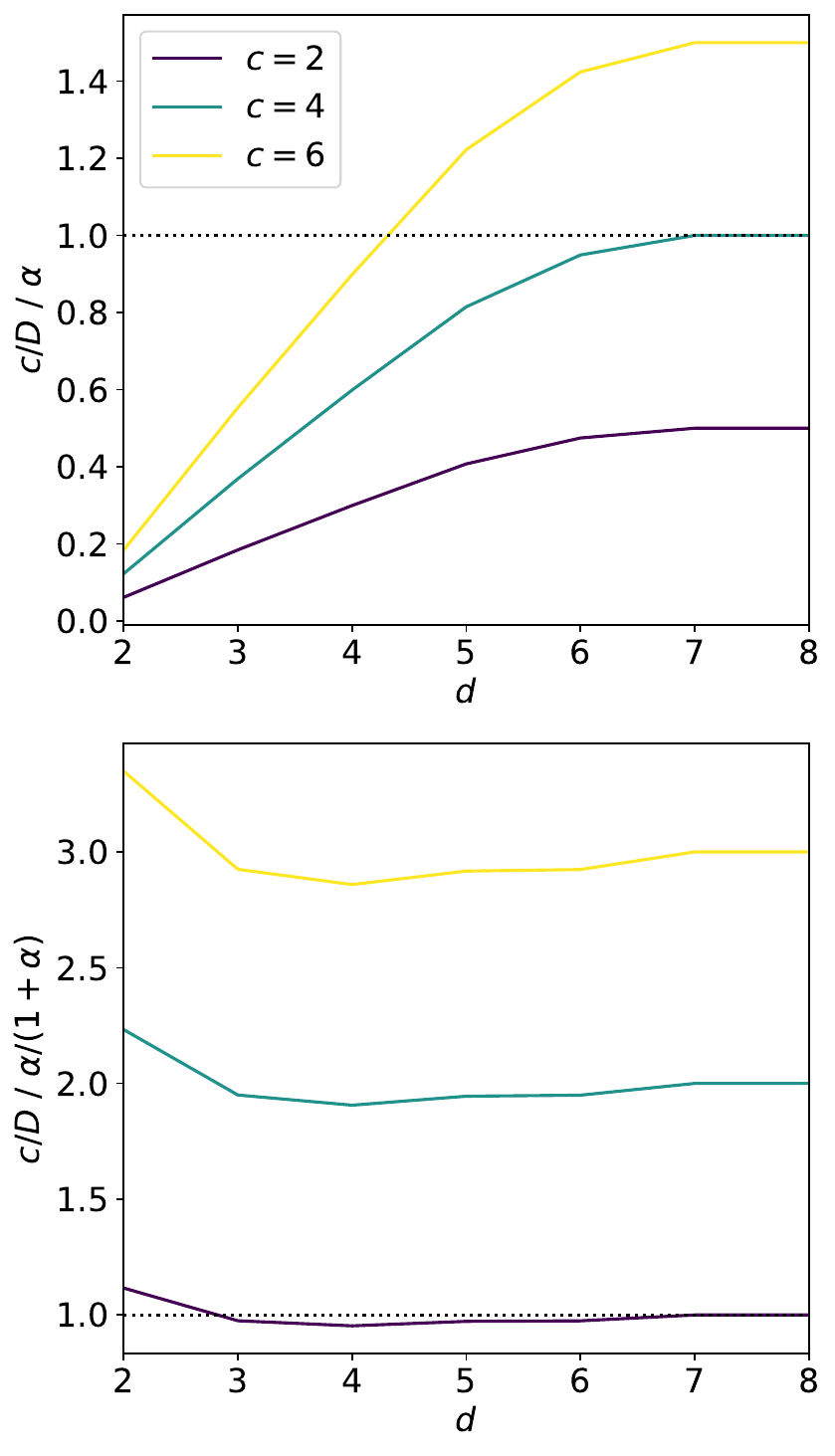}
    \caption{Exponent ratios controlling model scaling (top) and data scaling (bottom), meaningful when $p \lesssim p_c$. The overall scaling law regime transitions about a ratio of 1 for both.}
    \label{fig:scaling_comparison}
\end{figure}

Fig.~\ref{fig:scaling_comparison} shows the ratio of scaling exponents for several values of $c$, assuming that $p \lesssim p_c$. Model scaling begins to transition from manifold approximation to quanta scaling when $c = 4$. For data scaling, the transition begins at $c = 2$, suggesting that manifold approximation data scaling should commonly hold for natural datasets only when $p \gtrsim p_c$. In Appendix~\ref{appendix:scaling_law_theories}, we discuss in more detail how the proposed model's criticality regimes and limiting cases relate to previously proposed theories of neural scaling laws.

\section{Experiments}\label{sec:experiments}

We tested the derived model and data scaling laws in a minimal toy setting that allows efficient dataset generation and model fitting, with more general investigations left for future work. We trained a model based on nearest-neighbor regression on datasets generated from percolation simulations. Because such simulations are inherently random, we generated and fit multiple datasets\footnote{For all plots, unless noted otherwise, the random seed was 0.}. 

\subsection{Dataset}

Each dataset was made by simulating percolation clusters at criticality on a Bethe lattice. A Bethe lattice provides a suitable approximation for percolation on a high-dimensional lattice, and yields exact tree structure enabling efficient cluster generation and downstream computation. To emulate percolation on a $d$-dimensional lattice, we simulated a Bethe lattice with degree $z = 2d$ at the critical threshold, $p = p_c = 1/(z - 1) = 1/(2d - 1)$. We set $d = 100$.

Clusters were generated by breadth-first iteration starting from a root site. Each active site's number of neighbors was sampled from a binomial distribution $B(z -1,~p)$\footnote{Along with a site's parent, this gives a maximum of $z$ neighbors. For the root, $B(z,~p)$ was used instead.}. Active sites were added until all sites had neighbors or a maximum size threshold was reached. To ensure that all clusters had enough sites without using too much memory, clusters with fewer than $300$ or more than $3 \times 10^7$ sites were discarded. To avoid unduly distorting the distribution, datasets were limited to 316 clusters, chosen to yield one cluster above the maximum size threshold in expectation. For simplicity and efficiency, clusters were represented as graphs, without embedding them into a high-dimensional Euclidean space.

While simulating a cluster, real-valued target values were generated for each active site. This was done following a branching random walk, with each site's value randomly drawn from a normal distribution centered at its parent's value. Afterward, each completed cluster's values were standardized to have zero mean and unit standard deviation. Examples of simulated clusters with superimposed target functions are shown in Appendix~\ref{appendix:cluster_visualization}.

This generation procedure is intended to reflect our motivating assumptions in Sec.~\ref{sec:assumptions}. However, we do not formalize or instantiate them in an explicit target function in this work. Implicitly, a random walk represents a continuous target function, and the exclusive dependence of target values on cluster topology corresponds to general-purpose learning. Standardizing the target values eliminates noise from varying cluster loss scales, reducing experimental variance.

\subsection{Machine Learning Model}

We studied scaling laws using a model based on nearest-neighbor regression. This is a piecewise-constant nonparametric function approximator with an expected manifold approximation constant of $c = 2$. For all experiments, non-overlapping data subsets were randomly selected for training and testing. The number of training data points corresponds to both DOF (model scaling) and dataset size (data scaling). The loss function was mean squared error.

A key challenge was efficiently emulating an optimal model. We began with the simple and fast core of nearest-neighbor regression. The distance metric used was shortest-path graph distance, with every cluster's full graph representation provided and every data point's cluster identified. Because caching the distances between all site pairs consumed too much memory, nearest-neighbor values were efficiently computed by caching each site's distance to the root and an index to its parent, and exploiting the cluster's tree structure to reconstruct distances.

This model was augmented in two ways. First, a normal prior distribution was incorporated to provide less noisy predictions than naive nearest-neighbor regression. This was done using the Bayesian estimator, derived in Appendix~\ref{appendix:bayesian_estimate},

\begin{equation}
    y_\mathrm{pred} = \frac{y_\mathrm{nn}}{1 + \frac{d_\mathrm{nn}}{\sqrt{s}}},
\end{equation}

where $y_\mathrm{pred}$ is the prediction, $y_\mathrm{nn}$ the nearest neighbor value, $d_\mathrm{nn}$ the distance to the nearest neighbor, and $s$ the cluster size. When $d_\mathrm{nn}$ is large, $y_\mathrm{pred}$ approaches 0, the prior mean.

With this improved estimator, the scaling curves for individual clusters were found to be consistent with manifold approximation, $\mathcal{L} \sim N^{-c/D}$. Empirically, however, each cluster's scaling law had a different constant prefactor. We modeled this using cluster-dependent scale factors $m \ge 1$ to convert training data points $P$ to DOF $N$, i.e., $P = m N$. Because learning requires at least one DOF, this manifests in the scaling curve as an initial ``small data'' region or plateau of random-guess loss \citep{hestness2017deep}. 

Each cluster's value of $m$ was then fit using the scaling law, 

\begin{equation}\label{eq:parameterized_scaling_law}
    \mathcal{L} = \begin{cases}
    1,~P < m\\
    (P/m)^{-c/D},~\mathrm{otherwise}.
    \end{cases}
\end{equation}

Appendix~\ref{appendix:single_clusters} shows representative examples of fitted cluster scaling curves. Overall, $m$ varied between 1 and roughly 20, with an approximate central value of $m \sim 5$. For all following experiments, each cluster's assigned number of DOF was scaled up by its fitted $m$ to get the number of training points. Because $\alpha = 1$, $c = 2$, and $D = 4$ are fixed, the predicted scaling laws have no further free parameters.

\subsection{Results}

\begin{figure}
    \centering
    \includegraphics[width=\linewidth]{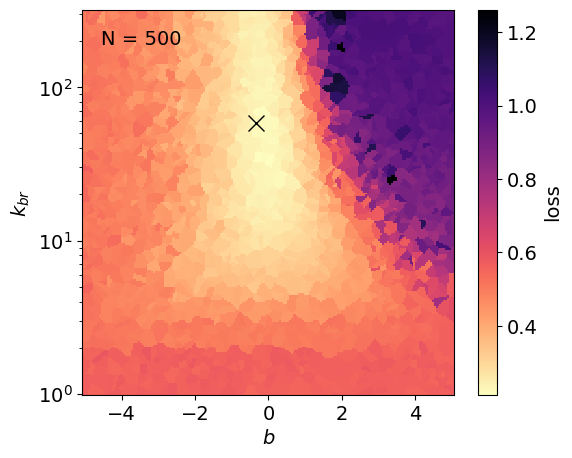}
    \caption{Loss achieved by a function approximator with parameterized DOF distribution for $N = 500$. A black cross indicates the parameters predicted to be optimal for model scaling.}
    \label{fig:model_scaling_loss}
\end{figure}

Eq.~\ref{eq:dof_allocation} gives the predicted optimal allocation of DOF among clusters for model scaling. We tested this using a parameterized DOF distribution with the functional form of Eq.~\ref{eq:dof_allocation}, but allowing $b$ and $k_\mathrm{br}$ to vary freely. The loss was computed for 2000 random combinations of $b$ and $k_\mathrm{br}$, with the results for $N = 500$ shown in Fig.~\ref{fig:model_scaling_loss}. Appendix~\ref{appendix:parameterized_model_loss} provides similar plots for other $N$. The best parameters appear consistent with the theoretically optimal ones.

\begin{figure}[ht]
    \centering
    \subfigure[Model scaling]{
        \includegraphics[width=0.95\linewidth]{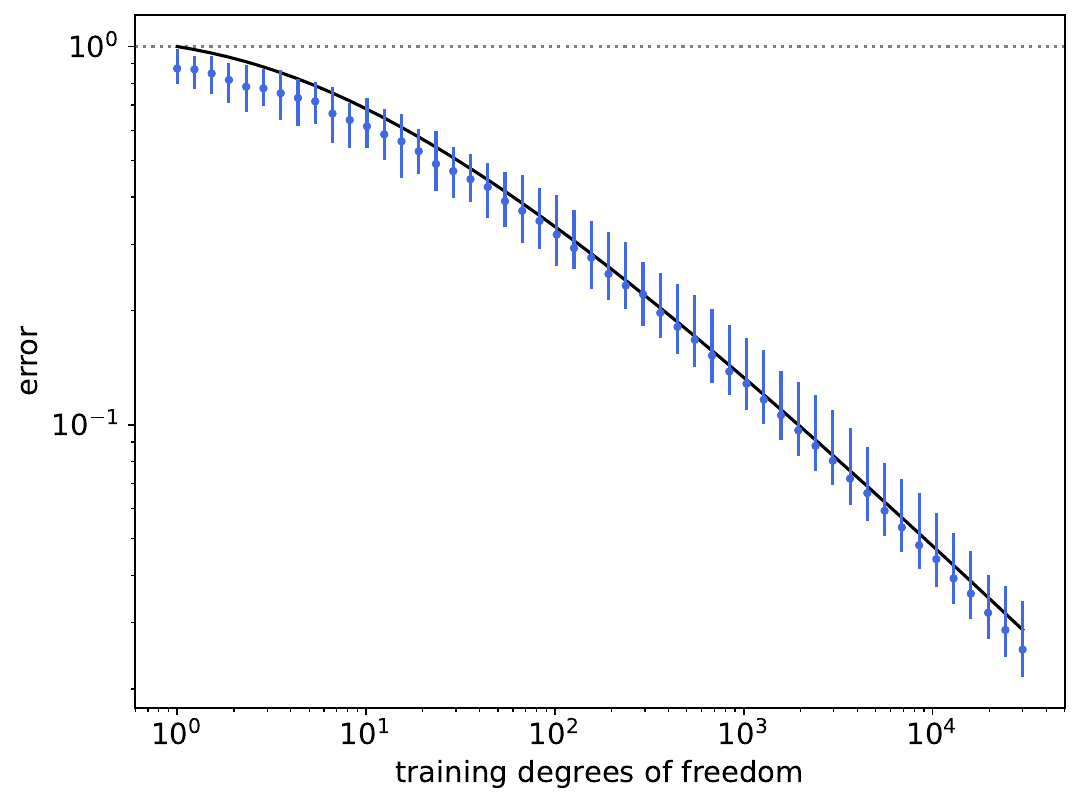}
    }
    \subfigure[Data scaling]{
        \includegraphics[width=0.95\linewidth]{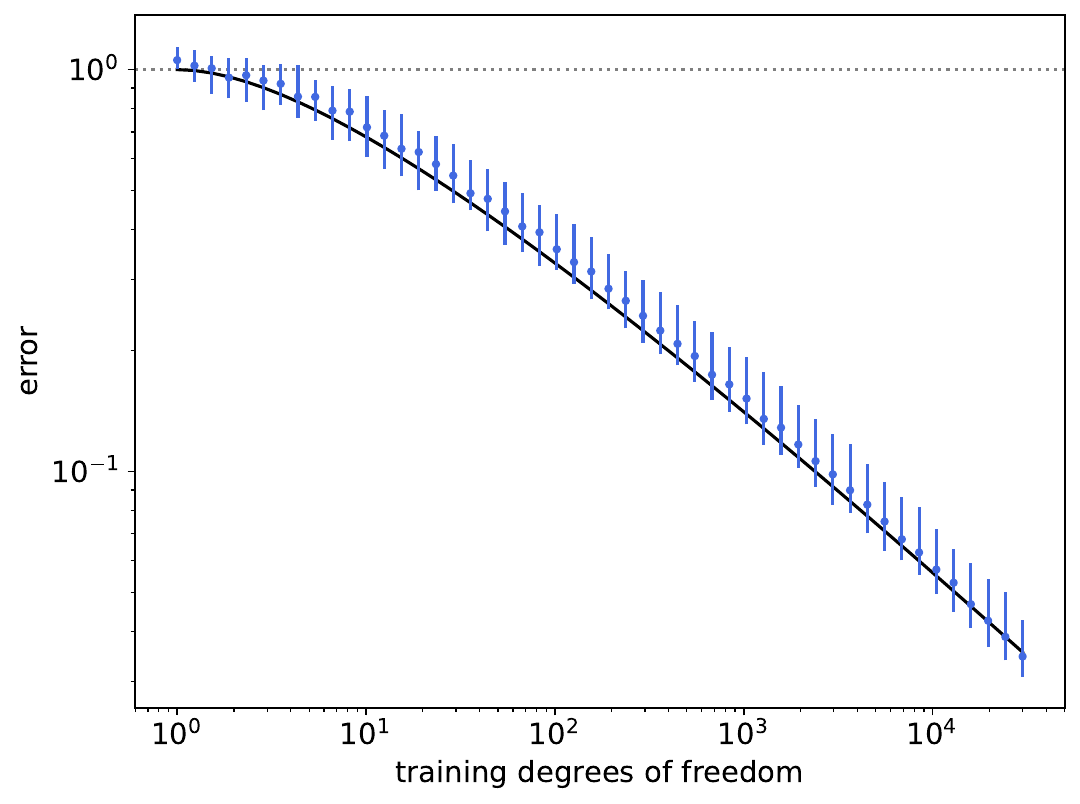}
    }
    \caption{Scaling laws computed using 50 random seeds. Results are shown as blue points with 1 standard deviation error bars. Theoretical predictions are overplotted in black (not a fit).}
    \label{fig:scaling_laws}
\end{figure}

Next, we investigated model and data scaling, with the DOF allocation for model scaling fixed to Eq.~\ref{eq:dof_allocation}. Scaling curves were computed using 50 random seeds to account for the randomly generated datasets' inherent variability. Fig.~\ref{fig:scaling_laws} shows the results. Each blue point with 1 standard deviation error bars shows the median with 16th and 84th percentiles. The results match theoretical predictions well.

\section{Discussion}\label{sec:discussion}

\subsection{Scaling Large Language Models}\label{sec:llm_scaling}

LLMs have achieved human-level performance across a bewildering array of tasks and disciplines \citep[e.g.][]{bubeck2023sparks}. This diversity suggests that subcritical percolation may model natural language well. Indeed, \citet{youn2016universal} used semantic network analysis across diverse human languages to reveal universal structure consisting of almost disconnected clusters of closely related concepts.

Understanding how close scaling laws achieved by LLMs\footnote{For LLM training, compute scaling can be derived from model and data scaling \citep{hoffmann2022training}, and this can be generalized to account for inference \cite{sardana2023beyond}.} are to theoretically optimal is essential for forecasting their future performance and transformative potential \citep[e.g.][]{branwen2020, cotra2020, epoch2023thedirectapproach}. We discuss the complex relation between scale and capabilities in Appendix~\ref{appendix:emergence}. Possibly the most influential measurement of LLM scaling laws comes from Chinchilla \citep{hoffmann2022training}, which fit the loss using the parametric form\footnote{The notation has been changed for compatibility with ours.},

\begin{equation}
    \mathcal{L}(P, \mathcal{D}) = E + \frac{A}{P^{\alpha_\mathrm{m,obs}}} + \frac{B}{\mathcal{D}^{\alpha_\mathrm{d,obs}}},
\end{equation}

with constants $A$, $B$, and $E$, and model and data scaling exponents $\alpha_\mathrm{m,obs}$ and $\alpha_\mathrm{d,obs}$. \citet{hoffmann2022training} reported exponents of $\alpha_\mathrm{m,obs} = 0.34$ and $\alpha_\mathrm{d,obs} = 0.28$. \citet{besiroglu2024chinchilla} conducted a statistical reanalysis to get exponents of $\alpha_\mathrm{m,obs} = 0.35 \pm 0.02$ and $\alpha_\mathrm{d,obs} = 0.37 \pm 0.02$.

As discussed in Appendix~\ref{appendix:dofandparameters},
a comparison to empirical model scaling requires converting DOF $N$ to parameters $P$. We look at two scenarios. First, the infinite-width relation $N = P$ yields $\alpha_\mathrm{m,th} = \alpha = 1$. Another conjecture could be $N \propto d_\mathrm{model}$. For example, associative memories in transformers \citep{geva2021transformer, meng2022locating} may yield scaling laws with respect to feedforward layer size $d_\mathrm{ff}$ \citep{cabannes2023scaling}, and generally $d_\mathrm{ff} \propto d_\mathrm{model}$. For Chinchilla, we estimate in Appendix~\ref{appendix:chinchilla} that $d_\mathrm{model} \propto P^{0.387}$, yielding $\alpha_\mathrm{m,th} = 0.387\cdot\alpha = 0.387$. Thus, more work is needed to tell if LLMs have approached our theory's limit to model scaling, or if more efficient scaling is possible.

For data scaling, we predict that $\alpha_\mathrm{d,th} = \alpha/(1 + \alpha) = 0.5$. Our theory therefore suggests that somewhat more efficient data scaling remains possible for LLMs.

\subsection{Data Distributions Near Criticality}

In this work, $p$ is a free parameter, and no inherent mechanism exists to push the data distribution to criticality \citep[cf. self-organized criticality,][]{bak1987self}. However, real-world datasets may in practice have near-critical $p$, because they're selected for feasible learning. A dataset with $p \ll p_c$ resembles random data and allows minimal generalization. Memorizing it is inefficient and useless. On the other hand, $p \gg p_c$ describes a dataset with simple functional structure but irreducible, strongly-coupled dependence on many input dimensions. It's very inefficient to learn, as the scaling exponent is proportional to $1/d$. Only near criticality with $p \approx p_c$ is learning both useful and efficient. In this regime, the data distribution contains one or more large but low-dimensional clusters in data space, corresponding to the well-known manifold hypothesis \citep{bengio2013representation}.

\subsection{Future Directions}

This work made theoretical predictions of neural scaling laws. It could be built on by training neural networks on toy datasets to test scaling regimes with $c > 2$; searching for percolation cluster structure in natural data distributions; and mechanistically interpreting neural networks to see whether they learn features that recapitulate dataset structure.

To derive power-law scaling, we assumed that data sampling was uniform and random. Non-uniform methods such as data pruning to undersample large clusters or oversample small ones \citep[e.g.][]{sorscher2022beyond}, or active learning to query for desired data \citep{ren2021survey}, could possibly outpace power-law data scaling. In addition, same-context samples can be correlated, so realistic sampling from a clustered data distribution may be nonergodic. A capacity tradeoff might then exist between modeling the data distribution and learning from context. Indeed, clustered data with many rare classes can drive emergent in-context learning in transformers \citep{chan2022data}. By linking neural networks' internal mechanisms to data distributional properties, we can better understand how these models learn and scale.

\section*{Acknowledgements}

Thanks to Philipp Kreer, Eduardo L\'opez, Eric Michaud, Adam Shai, and Alok Singh for feedback and discussions on drafts of this work. This research was supported by a grant from the Long-Term Future Fund (EA Funds). Research was sponsored by the National Aeronautics and Space Administration (NASA) through a contract with ORAU. The views and conclusions contained in this document are those of the authors and should not be interpreted as representing the official policies, either expressed or implied, of the National Aeronautics and Space Administration (NASA) or the U.S. Government. The U.S.Government is authorized to reproduce and distribute reprints for Government purposes notwithstanding any copyright notation herein.

\section*{Impact Statement}

This work theoretically investigates the origin of neural scaling laws. This study may advance the field of machine learning, which can have many potential societal consequences. Improved theoretical understanding of neural scaling laws may help in crafting policy aimed at ensuring that advanced machine learning systems are beneficial and safe.

\bibliography{main}
\bibliographystyle{icml2025}

\newpage
\appendix
\onecolumn

\section{General-Purpose Learning}\label{appendix:general_purpose_learning}

Our assumption of general-purpose learning describes a learning task that has latent data distributional structure independent of the format used to represent it. The latent structure comes from the world, while the input format is determined by the learning system. When this relationship exists, we equivalently say that the data's latent structure is \textit{extrinsic}, because no information about the external reality it derives from is built into the learning system. Indeed, deep neural networks trained on disparate datasets and modalities appear to learn convergent internal representations, consistent with independent general-purpose learning systems each modeling reality's shared extrinsic structure \citep{huh2024platonic}.

To illustrate, there are thousands of natural languages. These languages can have many observable realizations or forms, including speech, written words, visual gestures and signs, and token embedding vectors. Language in any form also has meaning: humans use it to communicate, which means it must in some way be structured by a correspondence with shared reality \citep{bender-koller-2020-climbing}. Language's form is essentially arbitrary, but its latent structure or meaning derives from external reality. A similar division between form and meaning exists in computer vision. A picture of a dog (say) is a meaningful image because it represents an actual or imagined dog, regardless of the values of its pixel-space representation.

Not all datasets have extrinsic structure. For example, a tabular dataset of handcrafted features doesn't, because its features define meaningful axes of variation by construction. The same is true of an algorithmic dataset representing a group-theoretic transformation or procedural task in which the symbolic forms are isomorphic with the mathematical operations represented \citep[e.g.][]{nanda2023progress, liu2022transformers, chughtai2023toy, michaud2024quantization}. In these cases, the input format provides built-in information relevant to the task at hand. 

General-purpose learning doesn't matter if the model fails to recognize or exploit a dataset's extrinsic structure. This might occur if adequate performance can be achieved by exclusively modeling surface statistics. Notably, \citet{bender2021dangers} claimed that language models are ``stochastic parrots'' that generate text based solely on statistical modeling of linguistic forms, without reference to meaning. By contrast, we assume that the model under consideration is not a stochastic parrot and can be usefully modeled as learning form-independent latent extrinsic structure.

A dataset's structure might be partly extrinsic and partly form-dependent, so that general-purpose learning is only partially realized. A language model may combine a meaningful world model with heuristics based on surface statistics. In addition, language can possess iconicity, or non-arbitrary resemblances between linguistic form and meaning \cite{perniss2014bridge}. We expect our theory to apply in proportion to the extent to which the dataset's structure is extrinsic.

\section{Expected Loss Reduction from Learning a Cluster Function}\label{appendix:loss_reduction}

In supervised learning, each in-distribution element of the data space constitutes a possible prediction task. There are then $pL^d$ possible tasks. Learning the function (or quantum) defined on a cluster of size $s$ reduces the error on $s$ of these tasks, so the expected loss reduction from learning an $s$-cluster's function is $\Delta \mathcal{L}_s \propto -s/(pL^d) \propto s$.

Neural networks are also commonly (pre-)trained on self-supervised generative objectives, including masked language modeling and next token prediction for language models, and masked denoising approaches in computer vision. In this setting, the model is trained to predict a portion of the input that has been masked or corrupted, by exploiting the surrounding context. The masked input may be one or more tokens or pixel patches, depending on the method and modality. If $m$ dimensions are masked, there are $\binom{d}{m} pL^d$ possible prediction tasks. As each prediction task is done independently in practice, we consider the case $m = 1$.

In this setting, the expected loss reduction from learning an $s$-cluster's function is given by the expected number of relevant prediction tasks multiplied by the expected loss reduction per relevant task, $\Delta \mathcal{L}_s = N_\mathrm{task} \times \Delta \mathcal{L}_\mathrm{task}$. If one dimension is masked, then the intersection of an $s$-cluster with the unmasked hypersurface has fractal dimension $D + (d - 1) - d = D - 1$, so that $N_\mathrm{task} \propto R_s^{D - 1}$.

The cluster's projection onto the masked dimension has expected length $R_s$. However, $\Delta \mathcal{L}_\mathrm{task}$ corresponds to how well learning a cluster function concentrates probability mass onto the correct output, which depends nonlinearly on $R_s$. If $R_s$ is small, the zero-information prior may overwhelm the cluster's contribution, leading to small $\Delta \mathcal{L}_\mathrm{task}$. If $R_s$ is large, the model can't pinpoint which of the cluster's possible output values is correct, and $\Delta \mathcal{L}_\mathrm{task}$ is again small. For moderately sized $R_s$, however, we will see below that $\Delta \mathcal{L}_\mathrm{task} \propto R_s$.

To do so, we consider the expected loss reduction from learning an $s$-cluster compared to a zero-information baseline for a prediction task with a cross-entropy loss function and vocabulary size $L$. We assume that the prediction task has exactly one correct output $i$, giving true labels $q_i = 1$, $q_{l \ne i} = 0$ for all $l \in (1, L)$. The zero-information baseline assigns an equal probability $p_l = p/\sum_{l = 1}^L p = 1/L$ to each possible output along the masked dimension, yielding loss $\mathcal{L} = -\log{\frac{1}{L}} = \log{L}$.

Now, suppose the model fully learns a relevant quantum corresponding to a cluster of size $s$ and radius $R_s$. We assume the cluster is small compared to the full data distribution, $s \ll pL^d$, so that learning it doesn't appreciably reduce the probability mass assigned to out-of-cluster elements. All data-space elements consistent with the unmasked input are possible outputs, with in-cluster elements being equally probable with unit weight and out-of-cluster elements having lower zero-information weight $p$. The probability that the correct output is in the cluster is then

\begin{equation}
    p_\mathrm{in} = \frac{R_s}{R_s + p(L - R_s)}.
\end{equation}

The probabilities that the model should predict for outputs in and out of the cluster, respectively, are

\begin{equation}
    p_{l,\mathrm{in}} = \frac{1}{R_s + p(L - R_s)},
\end{equation}

and

\begin{equation}
    p_{l,\mathrm{out}} = \frac{p}{R_s + p(L - R_s)}.
\end{equation}

The total expected loss reduction is, after some algebra,

\begin{align}\label{eq:l_task}
    \Delta \mathcal{L}_\mathrm{task} &= \mathcal{L}_\mathrm{in} + \mathcal{L}_\mathrm{out} - \mathcal{L}_\mathrm{baseline}\nonumber\\
    &= p_\mathrm{in}\left(-\log{p_{l,\mathrm{in}}}\right) + \left(1 - p_\mathrm{in}\right)\left(-\log{p_{l,\mathrm{out}}}\right) - \log{L}\nonumber\\
    &= \log{(1 + \delta)} + \frac{\delta}{1 + \delta}\frac{\log{p}}{1 - p},
\end{align}

where

$$
    \delta = \frac{(1 - p)R_s}{pL}.
$$

Note that Eq.~\ref{eq:l_task} vanishes either as $R_s \to 0$ or $R_s \to L$. We can think of $\delta$ as the ratio between the probability mass gained by the in-cluster elements when the cluster is learned and the total expected in-distribution probability mass. If we make the reasonable assumption that the probability mass attributable to learning a cluster is small compared to the expected total, then to first order,

\begin{equation}
    \Delta \mathcal{L}_\mathrm{task} \approx \delta + \delta\frac{\log{p}}{1 - p} = \frac{1 - p + \log{p}}{pL}R_s \propto R_s.
\end{equation}

The approximate expected loss reduction from learning an $s$-cluster is then $\Delta \mathcal{L}_s = N_\mathrm{task} \times \Delta \mathcal{L}_\mathrm{task} \propto R_s^{D - 1} R_s = s$.

\section{Neural Network Model Capacity}\label{appendix:dofandparameters}

Comparing theoretical predictions of model scaling with empirical measurements requires a way to consistently determine model capacity. Empirical scaling laws are typically reported in terms of the number of weight and bias parameters $P$, but relating parameter count to a notion of model capacity such as nonparameteric DOF $N$ isn't necessarily straightforward.

Many prior works theoretically investigating neural scaling laws appear to assume that model capacity and parameter count are directly proportional \citep[e.g.][]{sharma2022, bahri2021explaining, maloney2022solvable, michaud2024quantization}. But this assumption doesn't seem obviously true in general. For example, the capacity of a network with a narrow bottleneck layer (e.g. an autoencoder) is intuitively constrained by its bottleneck dimension, regardless of the other layers' sizes. The relation between parameter count and model capacity seems to be architecture-dependent. From an empirical standpoint, counting only non-embedding parameters provides significantly cleaner scaling laws \citep{kaplan2020scaling}. This suggests that model capacity may depend on how parameters are employed in a network, not just their total count. Also, regularization such as weight decay or dropout can reduce model capacity but isn't reflected in the raw parameter count.

Effective field theories of deep neural networks \citep{roberts2022principles} may provide a useful theoretical perspective. By analogy to statistical mechanics, one can describe a neural network either in terms of ``microscopic'' weight and bias parameters or ``macroscopic'' effective DOF. The macroscopic representation depends on architecture and scale. In the infinite-width limit, a deep neural network effectively interpolates the training data and is equivalent to a Gaussian processes \cite{nealbayesian, lee2018deep}. In this limit, a deep neural network's output can be described as depending on a number of fixed random macroscopic features equal to the number of parameters, giving $N = P$ \citep{jacot2018neural, lee2019wide}.

However, in practice, width and depth are typically scaled up in tandem \citep[e.g.][]{kaplan2020scaling, hoffmann2022training}. To analyze this regime, \citet{roberts2022principles} computed a perturbative expansion to the infinite-width limit, finding that representation learning arises when the network's depth-to-width ratio is small but finite\footnote{Representation learning can occur in the infinite-width limit if maximal-update parameterization is used \citep{yang2022tensor}.}. In this limit, the model output now depends on $P$ macroscopic features that are constrained by an implicit inductive bias to learn a nontrivial representation from the data. Since such a constraint implies that the learned features aren't independent, we conjecture that when representation learning occurs, $N < P$. Further theoretical and empirical investigation of model capacity in neural networks may be fruitful for shedding light on these issues.

\section{Model Scaling}\label{appendix:model_size_scaling}

We assume an idealized training procedure with unlimited training data and computation, and perfect optimization. We consider the pure scaling regime with irreducible error assumed to be negligible. Complexities resulting from relaxing these idealized assumptions could be empirically modeled using a flexible fitting procedure \citep[e.g.][]{hestness2017deep, rosenfeld2019constructive, alabdulmohsin2022revisiting, caballero2022broken}.

We consider a nonparametric model characterized by $N$ effective DOF. At convergence, the cluster of rank $k$ is allocated $n_k \ge 0$ DOF to model its function (quantum) such that

\begin{equation}\label{eq:n_normalization}
    N = \sum_{k=1}^{\infty} n_k.
\end{equation}

We assume that operations besides modeling cluster functions, such as identifying an input's corresponding cluster or converting the model's internal representations to the output format, consume negligible model capacity and can be neglected.

The total loss $\mathcal{L}$ is the sum of the loss from each cluster $k$,

\begin{equation}\label{eq:loss}
    \mathcal{L} = \sum_{k=1}^{\infty} \mathcal{L}_k,
\end{equation}

where each cluster's loss is given by manifold approximation \citep{sharma2022},

\begin{equation}\label{eq:parameter_L}
    \mathcal{L}_k = A_k k^{-(\alpha + 1)} \begin{cases}
        n_k^{-c_k/D_k} & n_k \geq 1\\
        1 & \mathrm{otherwise}.
    \end{cases}
\end{equation}

In Eq.~\ref{eq:parameter_L}, $A_k$ is a constant that determines the loss scale for cluster $k$. This constant combines numerical factors coming from sources including the efficiency of the model class and training procedure; the target label scale and any other factors in the loss function; the conversion between cluster rank $k$ and size $s$; and the function's complexity or rate of change. We assume that $A_k = A$ for all $k$, that is, all cluster functions have the same complexity. Without loss of generality we let $A = 1$. We further assume that all cluster functions are generic, giving $c_k = c$ for all $k$, where $c$ is a constant representing the complexity of the nonparametric function approximator. As discussed in Sec~\ref{sec:percolation_theory}, all clusters can be approximated as having the same intrinsic dimension $D$, so we let $D_k = D$ for all $k$. 

Each marginal DOF goes toward modeling the cluster function giving the greatest marginal loss reduction. Each cluster can be allocated one or more DOF, or none. In equilibrium, among clusters allocated DOF, the marginal loss reduction from adding DOF must be the same and equal to that of the first cluster allocated none, up to discretization. We call the rank of that cluster where a break occurs $k_\mathrm{br}$. Assigning DOF to any cluster after $k_\mathrm{br}$ would reduce the loss suboptimally.

In equilibrium, after training has converged, for any $k_1, k_2 < k_\mathrm{br}$, we have

\begin{equation}
    \frac{\partial}{\partial n_{k_1}}\left( n_{k_1}^{-c/D} k_1^{-(\alpha + 1)} \right) = \frac{\partial}{\partial n_{k_2}}\left( n_{k_2}^{-c/D} k_2^{-(\alpha + 1)} \right),
\end{equation}

where

\begin{equation}
    \frac{\partial}{\partial n_k}\left( n_k^{-c/D} k^{-(\alpha + 1)} \right) = -\frac{c}{D}n_k^{-(c/D + 1)}k^{-(\alpha + 1)},
\end{equation}

yielding

\begin{equation}
    \frac{n_{k_1}}{n_{k_2}} = \left( \frac{k_1}{k_2} \right)^{-\frac{\alpha + 1}{c/D + 1}}.
\end{equation}

To solve for $n_k$, we adopt the ansatz that, for $a = a(c, D, \alpha, N)$ and $b = b(c, D, \alpha, N) < 1$,

\begin{equation}
    n_k = \begin{cases} 
      ak^{b - 1} & k < k_\mathrm{br}\\
      0 & k \geq k_\mathrm{br}.
   \end{cases}
\end{equation}

Substituting, we obtain

\begin{equation}\label{eq:b}
    b = 1 - \frac{\alpha + 1}{c/D + 1} = \frac{c/D - \alpha}{c/D + 1}.
\end{equation}

In addition, we have from Eq.~\ref{eq:n_normalization},

$$N = \sum_{k=1}^{k_\mathrm{br}} ak^{b - 1} \approx \int_1^{k_\mathrm{br}} ak^{b - 1} dk = \frac{a}{b}\left(k_\mathrm{br}^b - 1\right),$$

giving

\begin{equation}\label{eq:a}
    a = \frac{bN}{k_\mathrm{br}^b - 1}.
\end{equation}

To solve for $k_\mathrm{br}$, we use the fact that the marginal loss reduction from allocating a marginal DOF for modeling any cluster with rank $k_i < k_\mathrm{br}$ is approximately equal to the marginal loss reduction from allocating the first DOF at $k_\mathrm{br}$,

\begin{equation*}
    -\frac{c}{D}\left( ak_i^{b - 1} \right)^{-(c/D + 1)} k_i^{-(\alpha + 1)} \approx -\frac{c}{D}\left( 1 \right)^{-(c/D + 1)} k_\mathrm{br}^{-(\alpha + 1)}.
\end{equation*}

Using the expressions for $b$ and $a$ from Eqs.~\ref{eq:b} and \ref{eq:a} and simplifying, we obtain

\begin{equation}\label{eq:kbr_eqn}
    bN = k_\mathrm{br}\left( 1 - k_\mathrm{br}^{-b} \right).
\end{equation}

If $c/D \gg \alpha$ and $c/D \gtrsim 1$, $b \approx 1$ and $k_\mathrm{br} \approx N + 1 \approx N$. If $c/D \approx \alpha$ or both $c/D \ll 1$ and $\alpha \ll 1$, $|b| \ll 1$ and Eq.~\ref{eq:kbr_eqn} approaches $N \approx k_\mathrm{br}\ln{k_\mathrm{br}}$. The solution is $k_\mathrm{br} = e^{W(N)} = N/W(N) \approx N/\ln(N/\ln{N})$, where $W(x)$ is the Lambert $W$ function. Finally, if $c/D \ll \alpha$ and $\alpha \gtrsim 1$, then $b < 0$ and $|b| > 1$, so $k_\mathrm{br} \approx \left(|b|N\right)^{1/(1 + |b|)}$. Summarizing,

\begin{equation}
    k_\mathrm{br} \approx \begin{cases}
        N & c/D \gg \alpha,~ c/D \gtrsim 1\\
        \left(\ln{\frac{N}{\ln{N}}}\right)^{-1}N & c/D \approx \alpha \mathrm{~or~} c/D \ll 1,~ \alpha \ll 1\\
        \left(|b|N\right)^{1/(1 + |b|)} & c/D \ll \alpha,~ \alpha \gtrsim 1.
    \end{cases}
\end{equation}

Putting everything together, the expected loss is given by

\begin{align}
    \mathcal{L} &= \sum_{k=1}^{k_\mathrm{br}} \left(ak^{b - 1}\right)^{-c/D}k^{-(\alpha + 1)} +  \sum_{k=k_\mathrm{br}}^{\infty} k^{-(\alpha + 1)}\nonumber\\
    &= \sum_{k=1}^{k_\mathrm{br}} a^{-c/D}k^{b - 1} + \sum_{k=k_\mathrm{br}}^{\infty} k^{-(\alpha + 1)}\nonumber\\
    &\approx \int_{1}^{k_\mathrm{br}} a^{-c/D}k^{b - 1} dk + \int_{k_\mathrm{br}}^{\infty} k^{-(\alpha + 1)} dk\nonumber\\
    &= \left(\frac{k_\mathrm{br}^b - 1}{b}\right)a^{-c/D} + \alpha^{-1}k_\mathrm{br}^{-\alpha}\nonumber\\
    &= \left(\frac{k_\mathrm{br}^b - 1}{b}\right)^{1 + c/D} N^{-c/D} + \alpha^{-1}k_\mathrm{br}^{-\alpha}\nonumber\\
    &= k_\mathrm{br}^{-(1 - b)(1 + c/D)}\left(\frac{k_\mathrm{br}\left( 1 - k_\mathrm{br}^{-b} \right)}{b}\right)^{1 + c/D}N^{-c/D} + \alpha^{-1}k_\mathrm{br}^{-\alpha}\nonumber\\
    &= \left(\frac{N}{k_\mathrm{br}} + \frac{1}{\alpha}\right)k_\mathrm{br}^{-\alpha}.
\end{align}

In the regime $c/D \gg \alpha$, $c/D \gtrsim 1$, $k_\mathrm{br} \approx N$, and we obtain

\begin{align}
    \mathcal{L} &\approx \left(1 + \frac{1}{\alpha}\right) N^{-\alpha}\nonumber\\
    &\propto N^{-\alpha}.
\end{align}

In the regime $c/D \ll \alpha$, $\alpha \gtrsim 1$, we obtain

\begin{align}
    \mathcal{L} &\approx \left(\frac{N}{ \left(|b|N\right)^{1/(1 + |b|)}} + \frac{1}{\alpha}\right) \left(|b|N\right)^{-\alpha/(1 + |b|)}\nonumber\\
    &= |b|^{-\frac{\alpha + 1}{1 + |b|}} \left(1 + \frac{1}{\alpha} 
|b|^{\frac{1}{1 + |b|}} N^{-\frac{|b|}{1 + |b|}} \right) N^{-\frac{\alpha - |b|}{1 + |b|}}\nonumber\\
    &= |b|^{-(c/D + 1)}  \left(1 + \frac{1}{\alpha} 
|b|^{\frac{1}{1 + |b|}} N^{-\frac{|b|}{1 + |b|}} \right) N^{-c/D}\nonumber\\
    &\approx \left(\frac{\alpha}{c/D + 1}\right)^{-(c/D + 1)}  \left(1 + \frac{1}{\alpha} N^{-\frac{\alpha}{1 + \alpha}} \right) N^{-c/D}\nonumber\\
    &\propto N^{-c/D}~\text{for large $N$}.
\end{align}

\section{Data Scaling}\label{appendix:data_scaling}

We assume that training examples are randomly sampled from the data distribution, which has $M \equiv pL^d$ elements. Suppose that the training set contains $\mathcal{D} \equiv \epsilon M$ examples, such that $\epsilon \ll 1$ and $\mathcal{D} \gg 1$. Then the cluster with rank $k$ will be sampled $l$ times, with $l$ having a binomial distribution,

\begin{equation}
    l \sim \binom{\mathcal{D}}{l}\left(1 - \alpha k^{-(\alpha + 1)}\right)^{\mathcal{D} - l} \left(\alpha k^{-(\alpha + 1)}\right)^l,
\end{equation}

where $\alpha k^{-(\alpha + 1)}$ is the probability that any given training example belongs to that cluster.

A cluster must be sampled at least once, $l \gtrsim 1$, to contribute to reducing the loss\footnotemark. If a cluster isn't sampled ($l = 0$), it can't be learned. For the cluster of rank $k$, the characteristic transition between these regimes occurs when $\mathcal{D} \alpha k^{-(\alpha + 1)} \approx 1$, yielding $k_\mathrm{br} \approx \left( \mathcal{D} \alpha\right)^{1/(1 + \alpha)}$.

\footnotetext{A cluster's size $s$ can be estimated if it's sampled multiple times, as the average distance between random cluster sites is proportional to $R_s$. If a cluster is only sampled once ($l = 1$), its overall size and shape can't be effectively estimated, but generalization is still possible in that sample's immediate neighborhood. One could model this using an estimated cluster size $s_\mathrm{est}(l)$, with $s_\mathrm{est}(l) \approx 1$ for $l = 1$ and $s_\mathrm{est}(l) \approx s$ for $l \gg 1$. We elide this complication as its effect on the total estimated loss is small.}

The total loss is then, up to an overall constant,

\begin{equation}
    \mathcal{L} \approx \sum_{k=1}^{k_\mathrm{br}} \sum_{l=0}^\infty p(l|k) l(k)^{-c/D} k^{-(\alpha + 1)} + \sum_{k=k_\mathrm{br}}^\infty k^{-(\alpha + 1)}.
\end{equation}

To simplify this expression, we'll replace the sums with integrals and make the further approximation that $p(l|k) \approx \delta(l(k) - \mathbb{E}(l|k))$, thereby substituting $l(k)$ with its mean value for each $k$. This yields

\begin{align}\label{eq:data_scaling_loss}
    \mathcal{L} &\approx \int_1^{k_\mathrm{br}} \left(\mathcal{D} \alpha k^{-(\alpha + 1)}\right)^{-c/D} k^{-(\alpha + 1)} dk + \int_{k_\mathrm{br}}^\infty k^{-(\alpha + 1)} dk\\\nonumber
    &= \frac{\left(\mathcal{D} \alpha\right)^{-c/D}}{\frac{c}{D}(1 + \alpha) - \alpha} \left( k_\mathrm{br}^{-\alpha + \frac{c}{D}(1 + \alpha)} - 1\right) + \alpha^{-1} k_\mathrm{br}^{-\alpha}\\\nonumber
    &= \frac{1}{\frac{c}{D}(1 + \alpha) - \alpha}\left(\left(\mathcal{D} \alpha\right)^{-\alpha/(1 + \alpha)} - \left(\mathcal{D} \alpha\right)^{-c/D}\right) + \alpha^{-1}\left(\mathcal{D} \alpha\right)^{-\alpha/(1 + \alpha)}\\\nonumber
    &= \frac{\alpha^{-\alpha/(1 + \alpha) - 1}}{1 - \frac{\alpha/(1 + \alpha)}{c/D}}\mathcal{D}^{-\alpha/(1 + \alpha)} + \frac{\alpha^{-c/D - 1}}{1 - \frac{c/D}{\alpha/(1 + \alpha)}}\mathcal{D}^{-c/D},
\end{align}

assuming that $c/D \ne \alpha/(1 + \alpha)$. In the limiting case where $c/D \gg \alpha/(1 + \alpha)$, the first term dominates and we obtain $\mathcal{L} \propto \mathcal{D}^{-\alpha/(1 + \alpha)}$. In the limiting case where $\alpha/(1 + \alpha) \gg c/D$, the second term dominates and we obtain $\mathcal{L} \propto \mathcal{D}^{-c/D}$.

If $c/D = \alpha/(1 + \alpha)$, we have instead of Eq.~\ref{eq:data_scaling_loss},

\begin{align}
    \mathcal{L} &\approx \left(\mathcal{D} \alpha\right)^{-c/D} \log{k_\mathrm{br}} + \alpha^{-1} k_\mathrm{br}^{-\alpha}\\\nonumber
    &= \alpha^{-\alpha/(1 + \alpha) - 1} \mathcal{D}^{-\alpha/(1 + \alpha)} + \alpha^{-c/D - 1}\frac{c}{D}\log(\alpha \mathcal{D}) \mathcal{D}^{-c/D}\\\nonumber
    &= \alpha^{-c/D - 1}\left(1 + \frac{c}{D}\log\alpha + \frac{c}{D}\log\mathcal{D} \right) \mathcal{D}^{-c/D}.
\end{align}

\clearpage
\section{Theories of Neural Scaling Laws}\label{appendix:scaling_law_theories}

The proposed neural scaling model encompasses several distinct criticality regimes and limiting cases that can be identified with previously proposed models of neural scaling laws. Table~\ref{tab:scaling_laws_theories} highlights several of these connections. In the subcritical regime, the scaling behavior depends on the relative magnitudes of $c/D$ and either $\alpha$ (for model scaling) or $\alpha/(1 + \alpha)$ (for data scaling). When $c/D$ is large, the achievable loss is bounded by the exponent of the cluster size distribution. This limit appears to correspond to the quantization model of neural scaling proposed by \citet{michaud2024quantization}. In both that work and this one, the predicted exponents for model and data scaling are $\alpha$ and $\alpha/(1 + \alpha)$, respectively.

\begin{table*}[ht]
    \centering
    \begin{tabular}{c|c|c|c}
        Criticality & Limit & Scaling & Related Works \\
        \hline
        \multirow{4}{*}{$p \le p_c$} & $c/D \gg \alpha$ & Model & \citet{michaud2024quantization} \\
                                     & $c/D \gg \alpha/(1 + \alpha)$ & Data & \citet{hutter2021learning, michaud2024quantization} \\
        \cline{2-4}
                                     & $c/D \ll \alpha$ & Model & \multirow{4}{*}{\citet{sharma2022, bahri2021explaining}} \\
                                     & $c/D \ll \alpha/(1 + \alpha)$ & Data &  \\
        \cline{1-3}
        \multirow{2}{*}{$p > p_c$} & \multirow{2}{*}{N/A} & Model & \\
                                   &                   & Data &  \\
    \end{tabular}
    \caption{Connections between the proposed model and related works.}
    \label{tab:scaling_laws_theories}
\end{table*}

Our model extends the work of \citet{michaud2024quantization} in several ways. We derive a power-law distribution of quantized subtasks from first principles, thus supporting the quantization hypothesis conjectured in that work. Furthermore, we quantitatively predict that $\alpha = 1$ (for $d \ge 6$), so $\alpha$ is not a free parameter. Qualitatively, our model suggests that neural networks learn quanta gradually in parallel, rather than fully one by one in sequence. This description complicates the connection between quantized subtasks and emergent capabilities, which we dicuss further in Appendix~\ref{appendix:emergence}. Unlike \citet{michaud2024quantization}, we don't consider training dynamics or make predictions for single-epoch scaling.

A data scaling exponent of $\alpha/(1 + \alpha)$ was also predicted by \citet{hutter2021learning} for a toy model of data scaling with Zipf-distributed data. \citet{hutter2021learning} obtained this exponent by considering a dataset of discrete labeled features and an algorithm that correctly predicts a feature if and only if it's been previously seen. The quantitative similarity between that model and ours can be understood by considering the discrete features to represent clusters and the learning algorithm to approximately correspond to the limit $c/D \to \infty$. \citet{cabannes2023scaling} obtained the same scaling behavior by analyzing the storage and retrieval of Zipf-distributed discrete associative memories using a simple model of a transformer layer.

In the supercritical regime, or in the subcritical regime when $c/D$ is small compared to either $\alpha$ (for model scaling) or $\alpha/(1 + \alpha)$ (for data scaling), the achievable loss is bounded by the model's ability to nonparametrically approximate the data distribution. This regime appears to correspond to the manifold approximation model proposed by \citet{sharma2022} and extended by \citet{bahri2021explaining}. Our theory extends these prior works by grounding the predicted scaling laws in a percolation model of the data distribution. A connected data distribution with well-defined intrinsic dimension emerges when $p > p_c$. When $p \le p_c$, $c/D$ scaling can still occur, with the data distribution consisting of disconnected clusters.

\citet{maloney2022solvable} proposed a solvable model of neural scaling laws in which the dataset is generated from a set of latent features with power-law spectral structure. The dataset's properties are controlled by two independent hyperparameters: the latent space dimension $M$, which must be larger than any other scale, and the spectral index $\alpha$, which determines the intrinsic dataset dimension such that $d_\mathrm{in} \propto 1/\alpha$. The dimensions $M$ and $d_\mathrm{in}$ represent different properties jointly characterizing the latent data manifold. In our model, the dataset is instead essentially characterized by a single hyperparameter, $p$ (since $D$ and $\alpha$ are fixed for $d \ge 6$). No explicit latent feature space is required. Instead, a power-law distribution of latent features --- clusters --- naturally emerges from a context-dependent target function and general-purpose learning.

\clearpage
\section{Visualizations of Simulated Clusters}\label{appendix:cluster_visualization}

\begin{figure}[h]
    \centering
    \subfigure[$s = 1995$]{
        \includegraphics[width=0.45\textwidth]{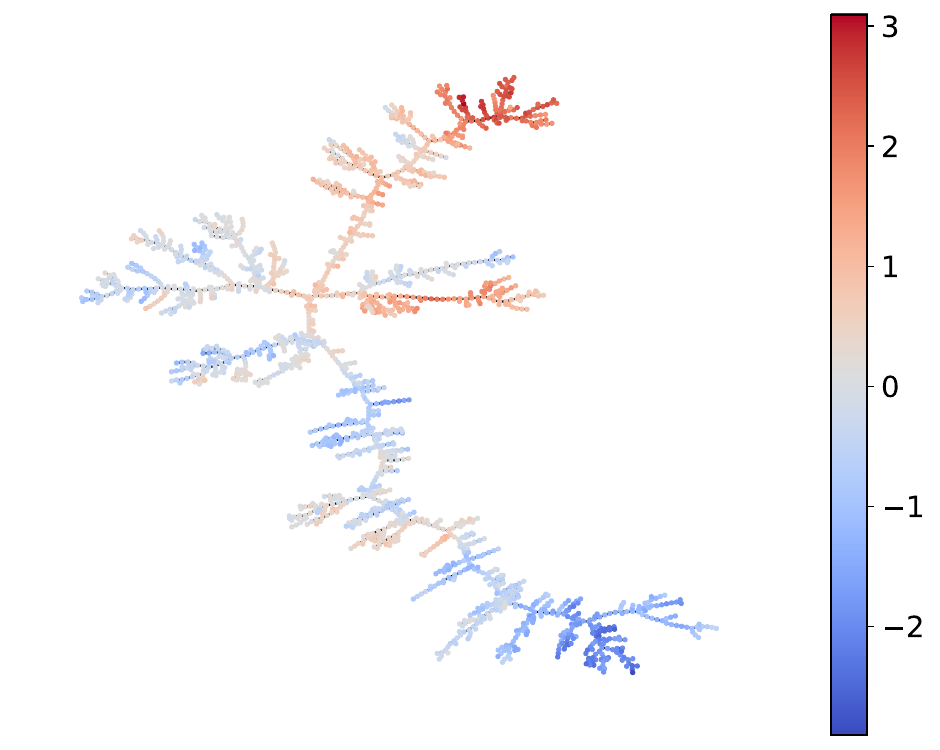}
    }
    \hfill
    \subfigure[$s = 1977$]{
        \includegraphics[width=0.45\textwidth]{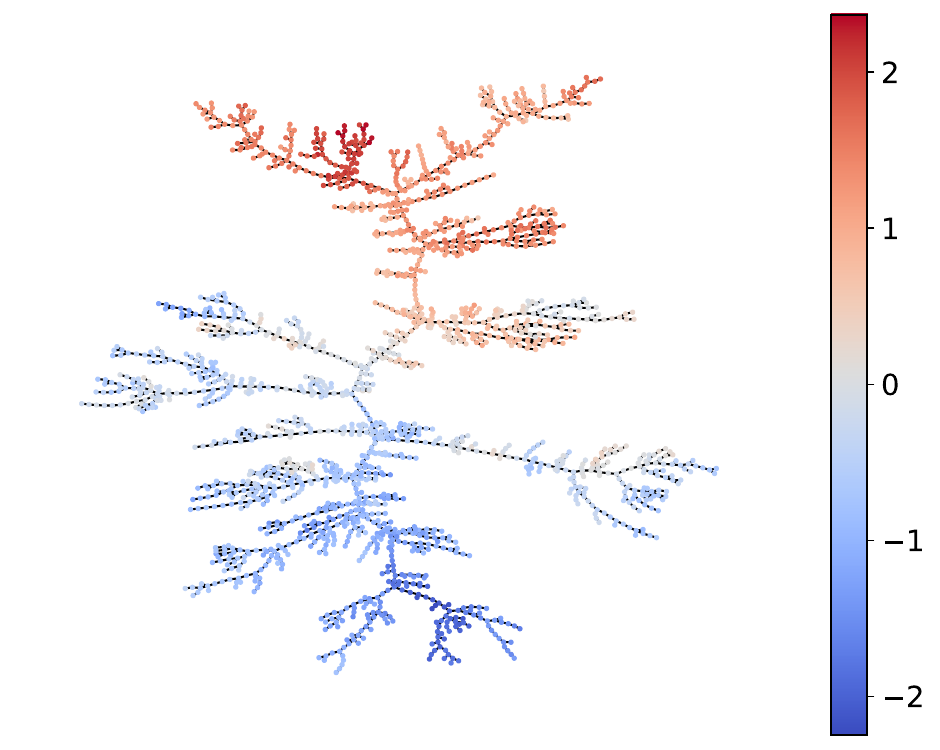}
    }
    
    \vspace{0.25cm}
    
    \subfigure[$s = 1963$]{
        \includegraphics[width=0.45\textwidth]{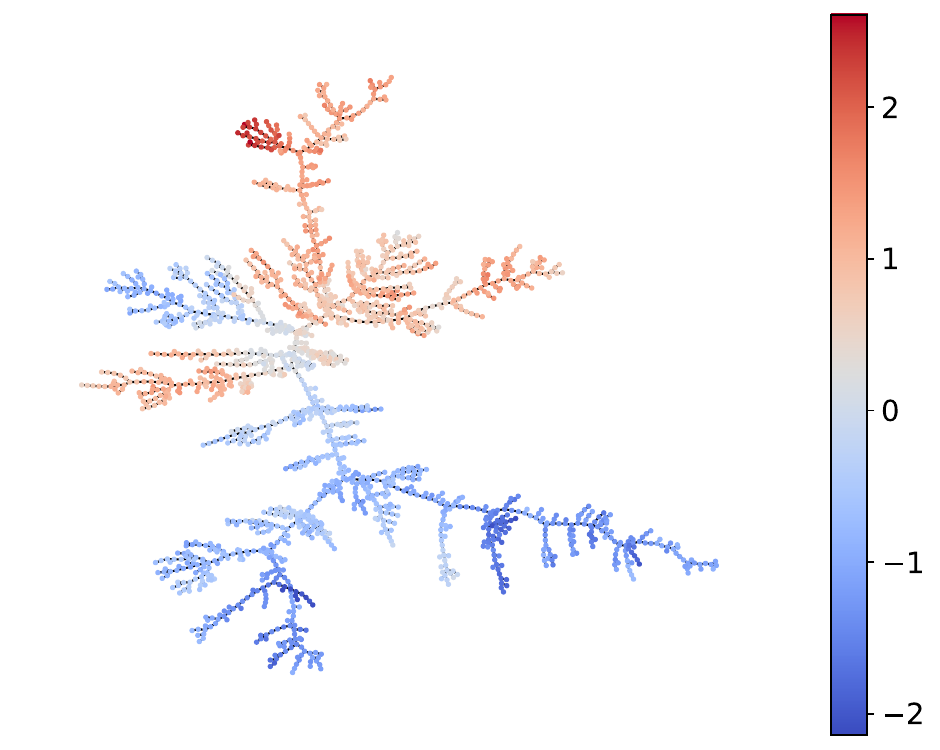}
    }
    \hfill
    \subfigure[$s = 1955$]{
        \includegraphics[width=0.45\textwidth]{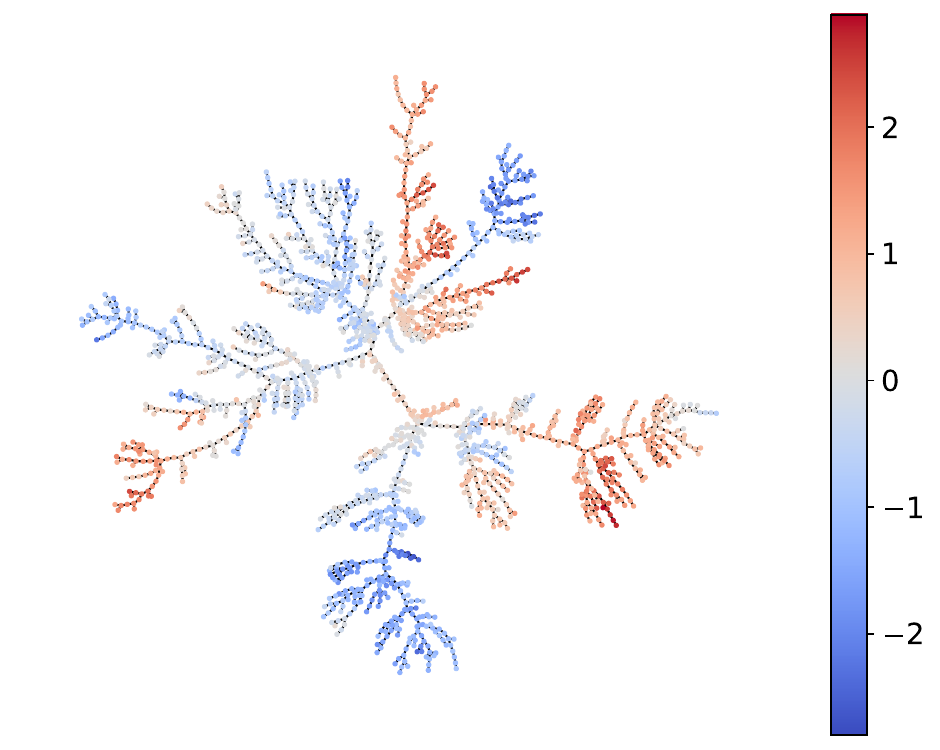}
    }
    \caption{Visualizations of percolation clusters simulated on a Bethe lattice. Site color indicates each cluster's superimposed function.}
    \label{fig:cluster_visualizations}
\end{figure}

Fig.~\ref{fig:cluster_visualizations} shows visualizations of functions defined on simulated percolation clusters. The clusters shown are restricted to less than 2000 sites to avoid incurring excessive computational cost for generating the visualizations.

\clearpage

\section{Bayesian Nearest Neighbor Estimator}
\label{appendix:bayesian_estimate}

The value predicted by a conventional nearest-neighbor estimator is the nearest neighbor's value, $y_\mathrm{pred} = y_\mathrm{nn}$. This estimator can be improved by using Bayesian inference to include prior information. In particular, adjusting the predicted value toward the prior mean reduces the expected error when the nearest neighbor is distant.

For our setup, the marginal distribution of sites' target values in a cluster can be modeled as standard normal, $y \sim \mathcal{N}(0, 1)$. We are given an observation of the nearest-neighbor training site's value $y_\mathrm{nn}$, and need to predict the value at a test site at a topological distance $d_\mathrm{nn}$, given by the number of sites along the minimum path length. The cluster size $s$ is known.

Each cluster's $y$ values were generated following a random walk with random terms $\epsilon$ initially distributed as standard normal. After standardization, the random terms have a distribution we model as $\epsilon \sim \mathcal{N}(0, \sigma_0)$, with a standard deviation $\sigma_0$ estimated below. The difference between any two sites' values is normally distributed, being a sum of normal terms. Since the random terms are independent, the standard deviation between sites at distance $d_\mathrm{rw}$ follows the relation,

\begin{equation}\label{eq:bethe_sigma}
    \sigma \sim \sigma_0 \sqrt{d_\mathrm{rw}}.
\end{equation}

The expression for Bayesian inference when the prior and likelihood function are both normal is well known \citep[see e.g.][Sec.~2.3]{bishop2006pattern}. In our case, it is given by

\begin{equation}\label{eq:bayesian}
    y_\mathrm{pred} = \frac{y_\mathrm{nn}}{1 + \sigma^2}.
\end{equation}

To estimate $\sigma$, we use fundamental properties of percolation on a Bethe lattice. First, the minimum path length $l$ between two sites on a percolation cluster is related to the geometric distance $R$ via an exponent $D_\mathrm{min}$ such that $l \propto R^{D_\mathrm{min}}$ \citep{stauffer1994introduction}. For a Bethe lattice, $D_\mathrm{min} = 2$. We recall that the size of a percolation cluster scales with the geometric length as $s \propto R^D$, where for a Bethe lattice $D = 4$. We then have the relation

\begin{equation}\label{eq:bethe_l}
    l \propto \sqrt{s}.
\end{equation}

Combining Eq.~\ref{eq:bethe_sigma} and Eq.~\ref{eq:bethe_l} to obtain a fixed marginal standard deviation with $d_\mathrm{rw} = l$ yields $\sigma_0 \propto s^{-1/4}$. Applying Eq.~\ref{eq:bethe_sigma} again with $d_\mathrm{rw} = d_\mathrm{nn}$, we obtain for the variance,

\begin{equation}\label{eq:variance}
    \sigma^2 = \mathrm{const}\times\frac{d_\mathrm{nn}}{\sqrt{s}}.
\end{equation}

Empirically, a constant of 1 works well. Combining Eq.~\ref{eq:bayesian} and Eq.~\ref{eq:variance} yields the estimator,

\begin{equation}
    y_\mathrm{pred} = \frac{y_\mathrm{nn}}{1 + \frac{d_\mathrm{nn}}{\sqrt{s}}}.
\end{equation}

\clearpage
\section{Scaling Laws for Individual Clusters}\label{appendix:single_clusters}

\begin{figure}[h]
    \centering
    \subfigure[Cluster $k = 0$]{
        \includegraphics[width=0.45\textwidth]{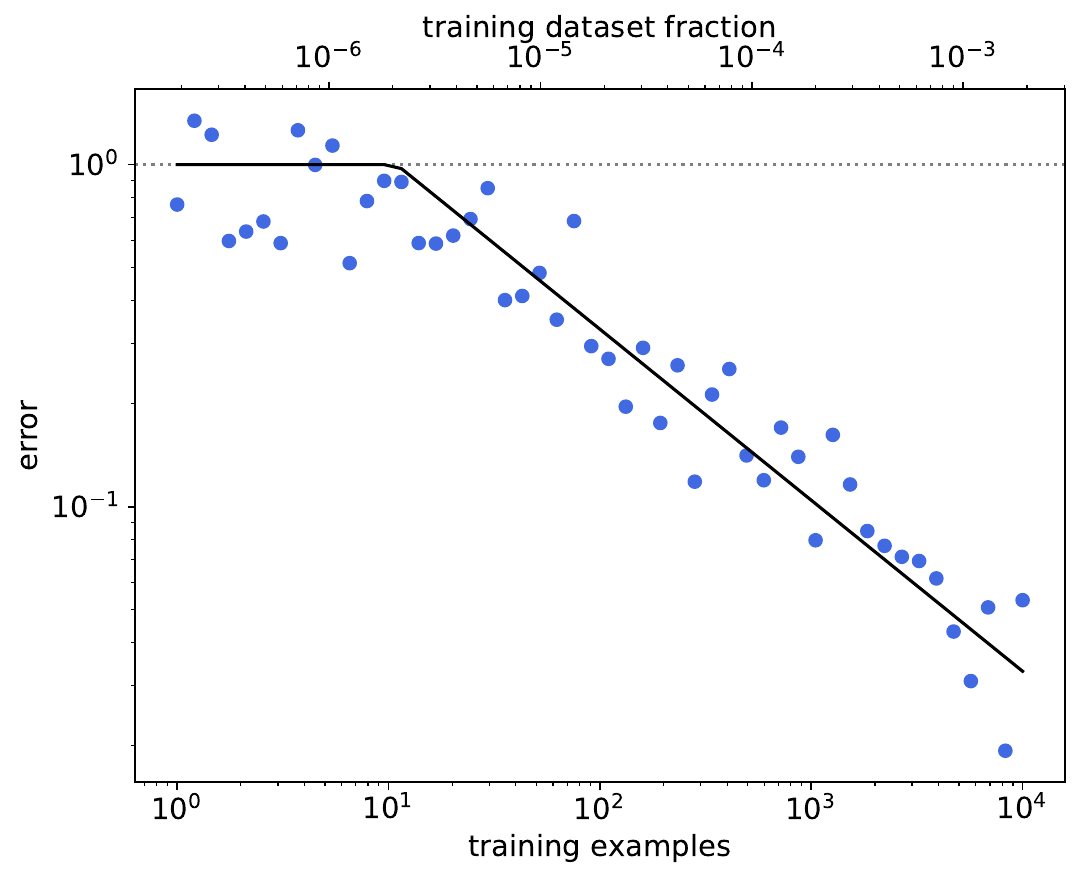}
    }
    \hfill
    \subfigure[Cluster $k = 1$]{
        \includegraphics[width=0.45\textwidth]{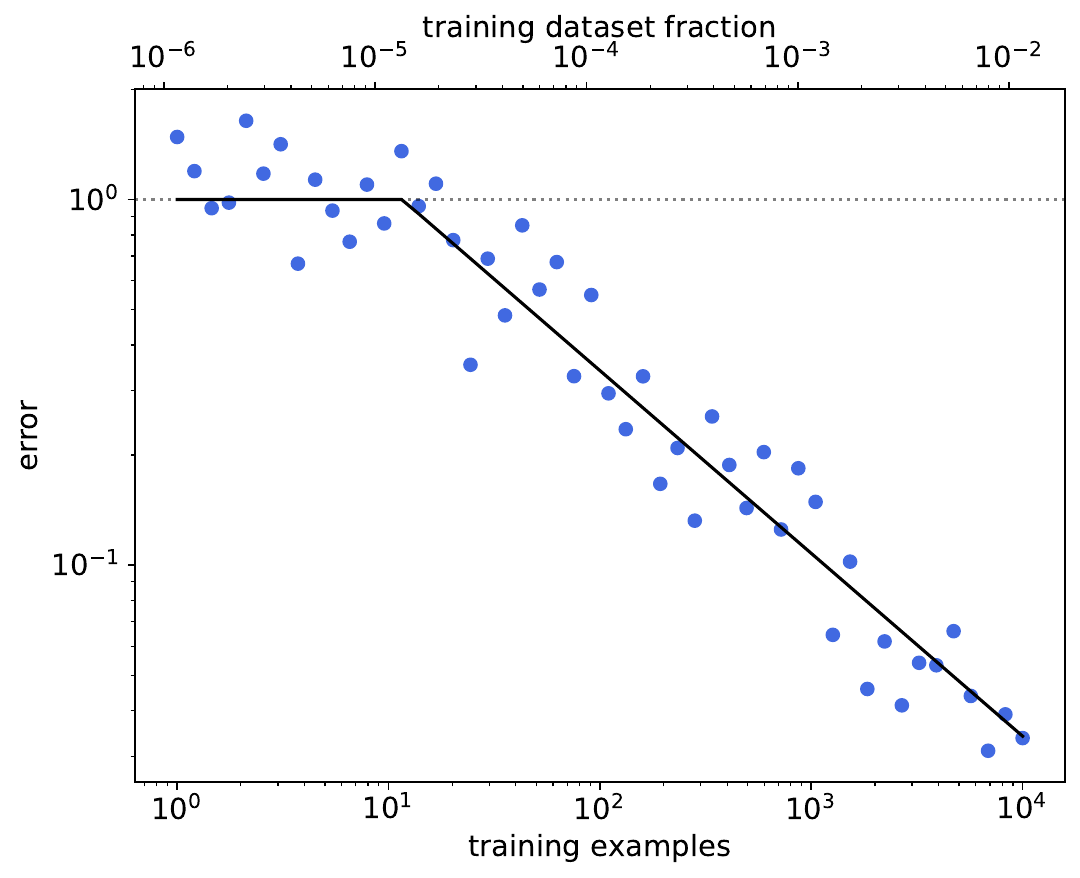}
    }
    
    \vspace{0.25cm}
    
    \subfigure[Cluster $k = 2$]{
        \includegraphics[width=0.45\textwidth]{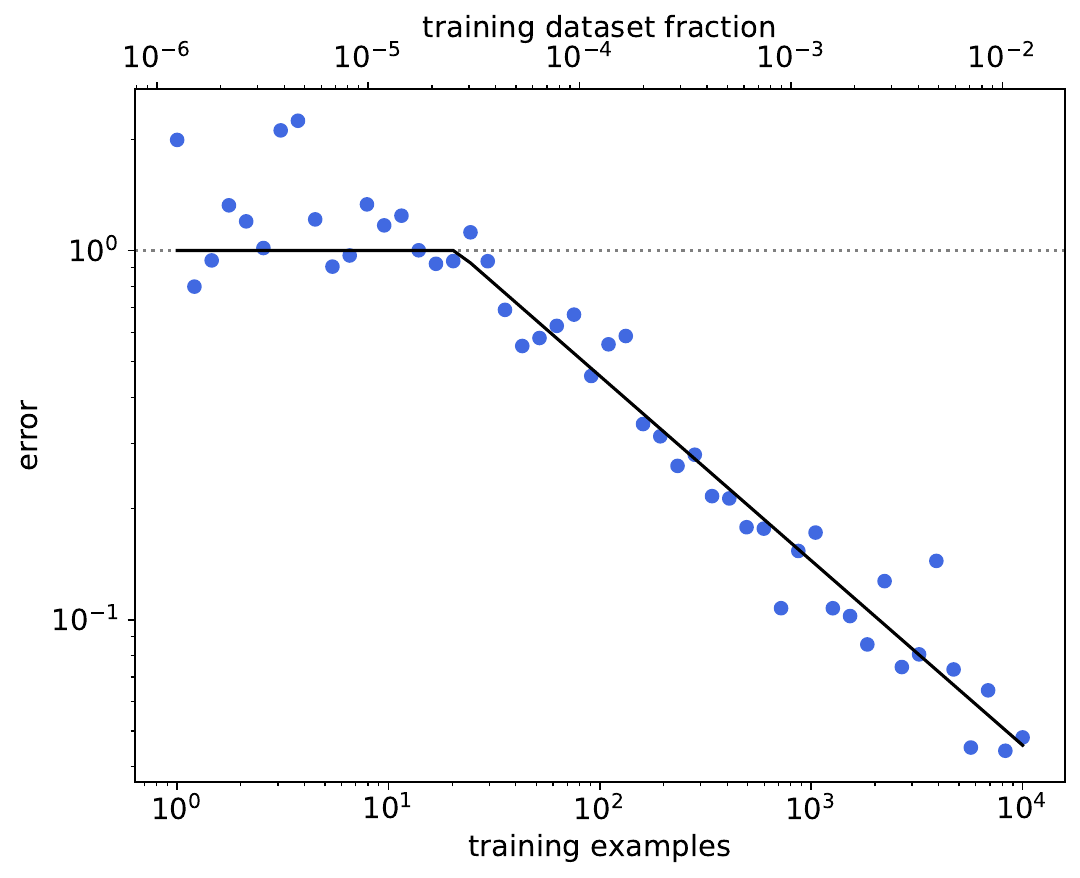}
    }
    \hfill
    \subfigure[Cluster $k = 315$]{
        \includegraphics[width=0.45\textwidth]{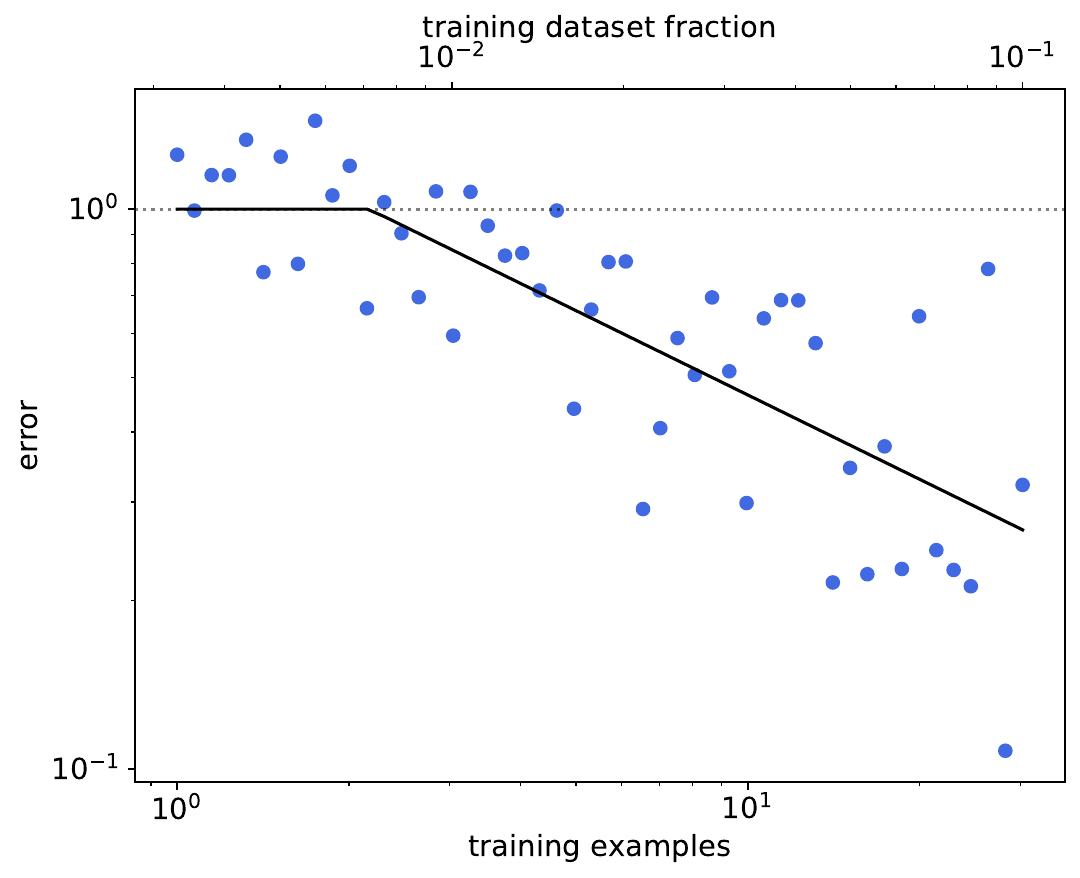}
    }
    \caption{Scaling curves for training and test sets restricted to individual clusters. Fits are shown in black, with the only free parameter for each being the break location. The first three plots show a dataset's largest three clusters, while the last shows its smallest cluster.}
    \label{fig:individual_cluster_scaling}
\end{figure}

Fig.~\ref{fig:individual_cluster_scaling} shows example scaling curves for individual clusters. After an initial break, each curve appears consistent with predicted manifold-approximation scaling, with power-law exponent $c/D = 0.5$.

\clearpage

\section{Parameterized DOF Allocation}\label{appendix:parameterized_model_loss}

\begin{figure}[h]
    \centering
    \subfigure[$N = 250$]{
        \includegraphics[width=0.4\textwidth]{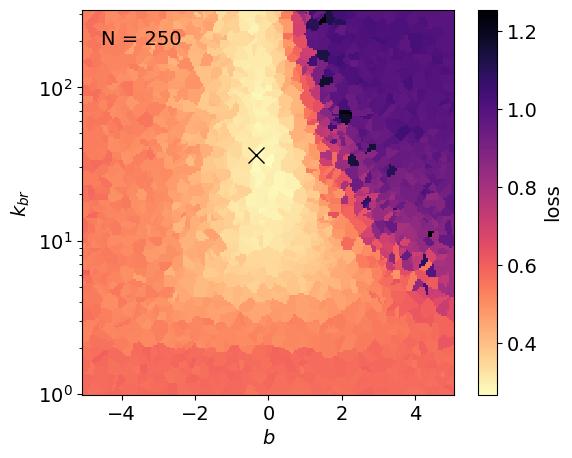}
    }
    \hfill
    \subfigure[$N = 1000$]{
        \includegraphics[width=0.4\textwidth]{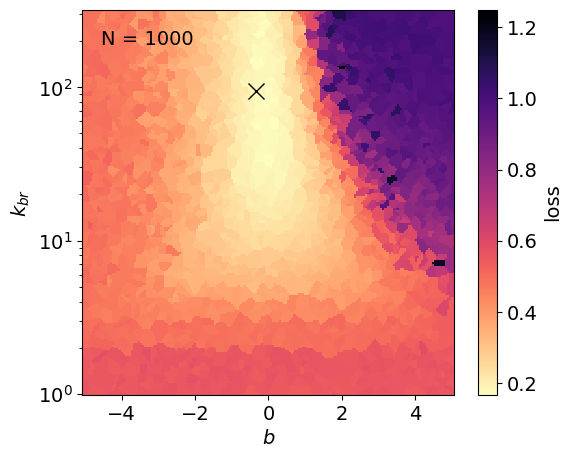}
    }
    \caption{Loss achieved by function approximators with parameterized DOF distributions, where $N$ is the total number of DOF. Black crosses indicate the parameters predicted to be optimal for model scaling.}
    \label{fig:model_scaling_loss_additional}
\end{figure}

Fig.~\ref{fig:model_scaling_loss_additional} shows the loss achieved with parameterized DOF allocations given different total values of DOF $N$. The case $N = 500$ is shown in the main text in Fig.~\ref{fig:model_scaling_loss}.

\section{Scaling Laws and Emergent Capabilities}\label{appendix:emergence}

In this work, we examined scaling laws relating a machine learning system's scale to its overall loss, but capabilities on given tasks are what determine utility and safety. Loss and capabilities can have a complex, nonlinear relationship. In particular, LLMs appear to display emergent capabilities\footnote{In this work, we also invoked ``emergence'' in an unrelated context to describe how power-law-distributed cluster structure appeared in the data distribution without being explicitly assumed.}, or abilities arising abruptly and unpredictably at particular model scales \citep{brown2020, wei2022emergent, srivastava2023beyond}. It remains unclear how best to reconcile discontinuous capability emergence with smooth power-law scaling laws. Indeed, \citet{schaeffer2023} suggested that apparent emergent capabilities may be artifacts of discontinuous metrics rather than fundamental model behavior changes. 

Several works have proposed phenomenological models to explain emergent capabilities. One approach models emergence through the composition of subtasks, since good task performance would emerge abruptly only once every subtask is learned \citep{okawa2023, arora2023theory}. Notably, \citet{lubana2024percolation} used percolation theory to model emergence in transformers trained on a formal language. \citet{lubana2024percolation} followed a very different approach than the one we employed, as they used percolation theory to model a network's learning dynamics rather than a static data distribution.

In addition, the quantization model of \citet{michaud2024quantization} yields a simple model of emergent capabilities, since a capability involving only one quantum would appear abruptly at the scale that it's learned. \citet{michaud2024quantization}, borrowing genetics terms, called a prediction problem ``monogenic'' if it involves only one quantum and ``polygenic'' if influenced by multiple quanta. They examined model scaling curves for individual natural language prediction tasks, finding mostly smooth scaling curves, but with some exhibiting abrupt performance improvements at particular scales. \citet{michaud2024quantization} identified smooth scaling curves with polygenic behaviors and abrupt ones with monogenic behaviors.

Our theory's subcritical regime broadly accords with \citet{michaud2024quantization}, but with additional complicating considerations. First, polygenic behavior occurs if a prediction problem is compatible with multiple subtasks due to noisy, masked, or ambiguous input, making the optimal behavior a mixture distribution. Second, because many subtasks are learned in parallel, both monogenic and polygenic prediction problems can manifest smooth scaling curves. Third, performance on a problem can indeed improve abruptly at the smallest model size a relevant quantum is learned. However, cluster functions are modeled smoothly, so even monogenic problems might not yield noticeable discontinuities. Finally, violations of general-purpose learning (e.g. exploitation of memorized surface-level patterns) have undefined scaling behavior, possibly compatible with abrupt performance improvements.

\section{Chinchilla Scaling Law}\label{appendix:chinchilla}

\begin{figure}[h]
    \centering
    \includegraphics[width=0.5\linewidth]{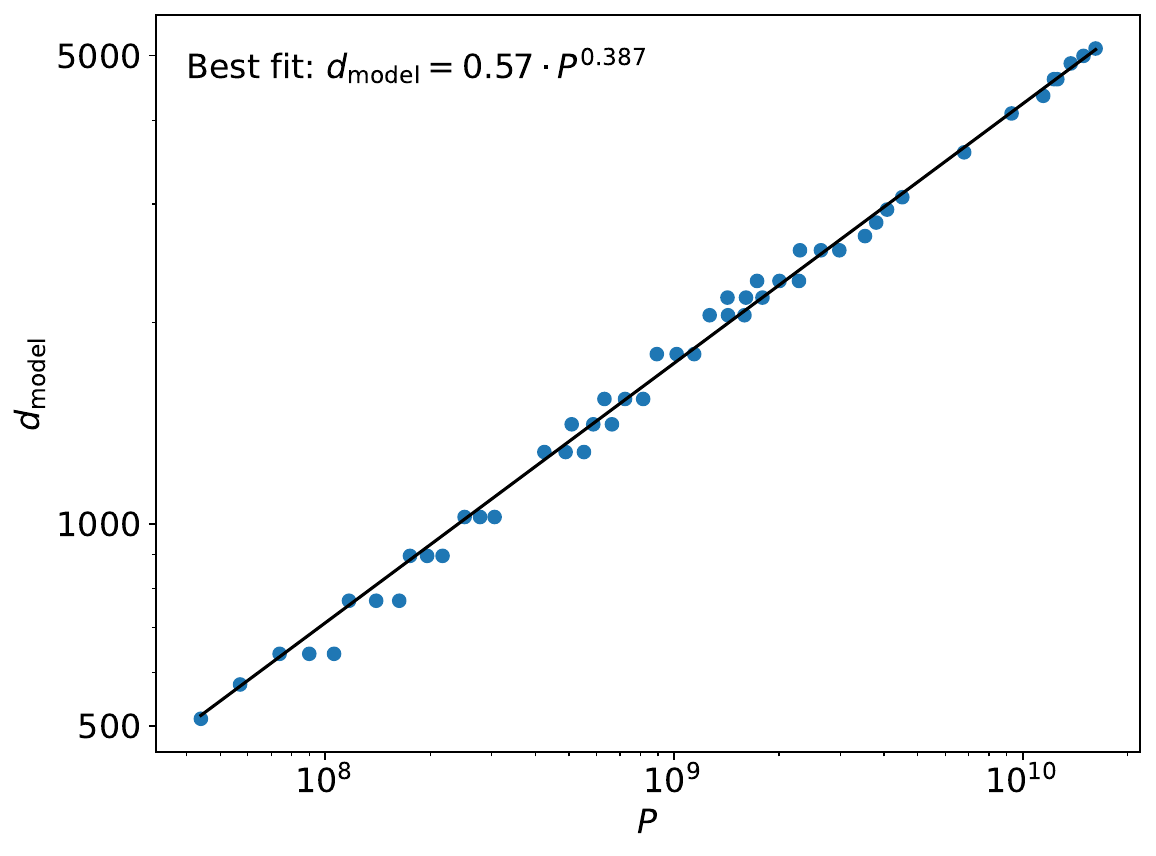}
    \caption{Relation between model width $d_\mathrm{model}$ and parameters $P$ for various sizes of the Chinchilla model family. Data points are those reported by \citet[][Table A9]{hoffmann2022training}.}
    \label{fig:chinchilla_scaling}
\end{figure}

Fig.~\ref{fig:chinchilla_scaling} shows the relationship between model width and number of parameters in the Chinchilla model family, using the architectural hyperparameters reported by \citet[][Table A9]{hoffmann2022training}. We fit the parametric relation $d_\mathrm{model} = CP^\beta$, obtaining best-fit parameters $C = 0.57 \pm 0.05$ and $\beta = 0.387 \pm 0.004$.

\end{document}